\documentclass[10pt,twocolumn,twoside]{IEEEtran}
\usepackage{hyperref}
\usepackage[dvipsnames]{xcolor}
\usepackage{amssymb}
\usepackage{caption}
\usepackage{amsfonts}
\usepackage{booktabs}
\usepackage{bm}
\usepackage{float}
\usepackage{algorithm}
\usepackage{algorithmic}
\usepackage{multicol}
\usepackage{multirow}
\usepackage{diagbox}
\usepackage{adjustbox}
\makeatletter

\usepackage{graphicx}
\usepackage{subfigure}
\makeatletter
\def\maketag@@@#1{\hbox{\m@th\normalfont#1$^*$}}
\makeatother

\newcommand{\eg}{\textit{e.g.}}
\newcommand{\etal}{\textit{et al.}}

\newcommand{\ie}{\textit{i.e.}}

\newcommand{\DJS}{D_{{JS}}\hspace{-1pt}}

\usepackage{newfloat}
\usepackage{listings}
\floatstyle{ruled}
\newfloat{listing}{tb}{lst}{}
\floatname{listing}{Listing}
\usepackage{lipsum}
\usepackage[switch]{lineno}
\ifCLASSINFOpdf
\else
\fi
\begin{document}
\title{FS-BAN: Born-Again Networks for \\ Domain Generalization Few-Shot Classification}

 \author{Yunqing Zhao
 \and
 {\hspace{3mm}}
 Ngai-Man Cheung

 \thanks{Yunqing Zhao and Ngai-Man Cheung are with the Information Systems Technology and Design Pillar, Singapore University of Technology and Design, Singapore 487372
 (email:
 yunqing\_zhao@mymail.sutd.edu.sg,
 ngaiman\_cheung@sutd.edu.sg).
 Correspondence to: Ngai-Man Cheung.
 }
 }


\markboth{
IEEE TRANSACTIONS ON IMAGE PROCESSING, 2023
}%
{
ZHAO \MakeLowercase{\textit{et al.}}: 
FS-BAN: Born-Again Networks for Domain Generalization Few-Shot Classification}



\maketitle

% % As a general rule, do not put math, special symbols or citations
% % in the abstract or keywords.
% \begin{abstract}
% The abstract goes here.
% \end{abstract}

% % Note that keywords are not normally used for peerreview papers.
% \begin{IEEEkeywords}
% IEEE, IEEEtran, journal, \LaTeX, paper, template.
% \end{IEEEkeywords}

% \fi
\IEEEpeerreviewmaketitle

% Abstract

\begin{abstract}
    Conventional Few-shot classification (FSC) aims to recognize samples from novel classes given limited labeled data. 
    Recently, domain generalization FSC (DG-FSC) has been proposed with the goal to recognize novel class samples from unseen domains.
    DG-FSC poses considerable challenges to many models due to the domain shift between base classes (used in training) and novel classes (encountered in evaluation).
    In this work, we make two novel contributions to tackle DG-FSC.
    Our first contribution is to propose Born-Again Network (BAN) episodic training and comprehensively investigate its effectiveness for DG-FSC.
    As a specific form of knowledge distillation, BAN has been shown to achieve improved generalization in conventional supervised classification with a closed-set setup. This improved generalization motivates us to study BAN for DG-FSC, and we show that BAN is promising to address the domain shift encountered in DG-FSC.
    %can also effectively improve the recognition of novel classes from unseen domain in few-shot setup. 
    Building on the encouraging findings, our second (major) contribution is to propose Few-Shot BAN (FS-BAN), a novel BAN approach for DG-FSC. Our proposed FS-BAN includes novel multi-task learning objectives: Mutual Regularization, Mismatched Teacher, and Meta-Control Temperature, each of these is specifically designed to overcome
    central and unique challenges in DG-FSC, namely overfitting and domain discrepancy. 
    We analyze different design choices of these techniques.
    We conduct comprehensive quantitative and qualitative analysis and evaluation over six datasets and three baseline models. The results suggest that our proposed FS-BAN consistently improves the generalization performance of baseline models and achieves state-of-the-art accuracy for DG-FSC.
    {
     Project Page:
     \href{https://yunqing-me.github.io/Born-Again-FS/}{{\color{RubineRed}{yunqing-me.github.io/Born-Again-FS/}}}.
    }
\end{abstract}

\begin{IEEEkeywords}
    Few-shot classification, domain generalization, born-again network, episodic training, meta-learning.
\end{IEEEkeywords}

% introduction
\section{Introduction}
 \label{sec1}
 
\IEEEPARstart{W}{hile} 
 modern deep learning models achieve superior performance in many visual recognition tasks, \eg, image classification \cite{deng2009imagenet}
 and object detection \cite{redmon2016yolo}, they require a large number of labeled data during training \cite{sun2017revisiting_google}.
 In contrast, in few-shot classification (FSC)  \cite{fei2006one, snell2017prototypical, sung2018relationnet, finn2017maml},
 the models are required to classify samples from \textit{novel} categories given only \textit{a few} labeled data from each category.
\begin{figure}[t]
    \centering
    \scriptsize
    {
        \centering
        \includegraphics[width=0.485\textwidth]{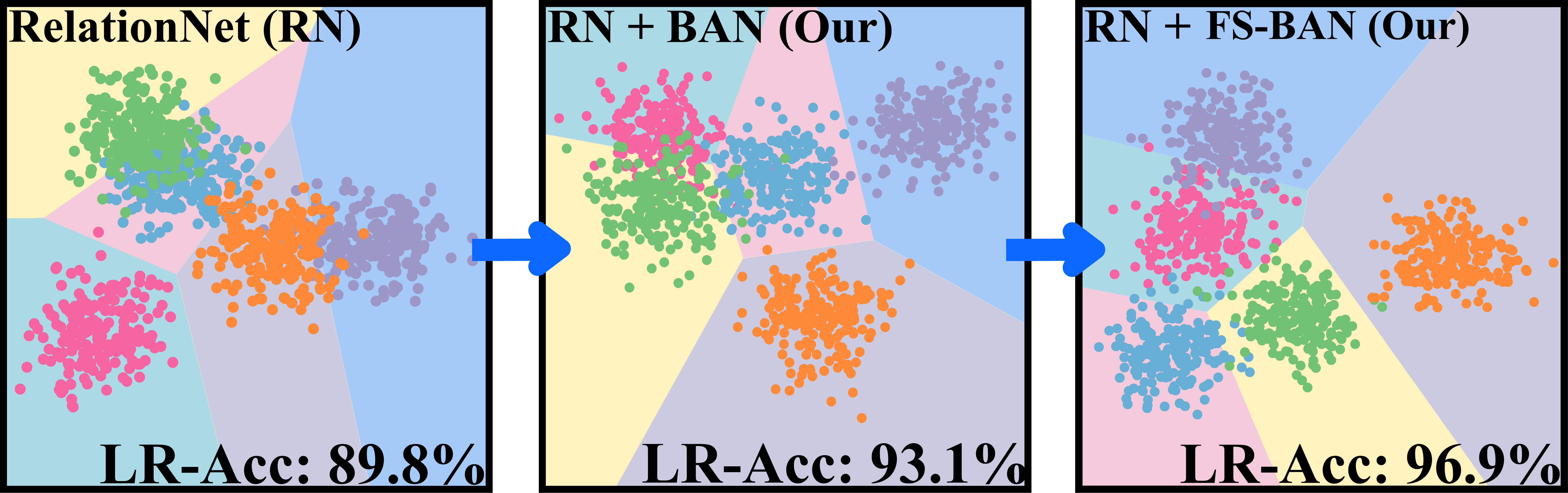}
    }
    \vspace{-4 mm}
    \caption{
    In this visualization, we select 5 novel classes with 200 query samples per class from an unseen domain (Places \cite{zhou2017places}). Each point indicates the feature representation of
    RelationNet (RN)
    \cite{sung2018relationnet} with 
    backbone network (ResNet-10 \cite{he2016resnet}), projected by LDA \cite{gold-icml2020-unraveling, mika1999fisher}. We use the linear regression prediction accuracy (``LR-Acc'') to demonstrate the improved decision boundaries: baseline RN (left), BAN episodic training applied to RN (mid), and our further proposed FS-BAN applied to RN (right). See numerical results and comparisons in Sec. \ref{sec5}.
    }
    \vspace{-5 mm}
    \label{fig1}
 \end{figure}
 
  \subsection{Domain Generalization FSC}
  Recently, meta-learning based FSC  \cite{sung2018relationnet, vinyals2016matching, finn2017maml, lee2019meta-opt}
  has achieved outstanding performance in the {\em single} domain setup, where the base classes for training and the novel classes for evaluation are from the same domain. However, in real-world applications, the deployed models are often required to classify objects from domains that are unseen during training, given limited labeled data (\eg, recognize rare bird species in a fine-grained setup \cite{tseng2020cross}).  In particular, our work addresses this challenging \textit{domain generalization} (DG) FSC: to recognize samples from novel classes of unseen domains given only a few labeled data of each class. 
  We follow recent DG-FSC works \cite{tseng2020cross,sun2020explanation} and assume to have several seen domains during training; however, we do not have access to samples from the unseen domains which will be encountered during evaluation. 
  DG-FSC has attracted a fair amount of attention recently \cite{chen2019closer, tseng2020cross,sun2020explanation, liu2020negative-margin}.
  Due to the significant discrepancy between the seen domains used in training and the unseen domains encountered in evaluation, existing FSC models designed only for the single domain setup often perform poorly \cite{chen2019closer}. Therefore, DG-FSC still has much room for improvement in generalization under the domain shift setup.

 \subsection{Born-Again Networks (BANs)}
 In their pioneer work, Breiman and Shang \cite{Breiman:1996} proposed {\em born-again trees}. 
 Given a complex predictor, \eg, a model with multiple trees ensemble, they train a single tree which outputs (decisions) match that of the complex predictor. This single born-again tree is simple and more interpretable compared to the complex predictor while it still maintains a decent decision performance \cite{vidal2020born-again-tree-ensemble}.
 More recently,
 \cite{furlanello2018bornagain}
 investigated the knowledge transfer
 \cite{hinton-distill}
 from one model (the teacher) to another model (the student).
 Focusing on the conventional image classification tasks, they first train the teacher network to convergence using the standard cross-entropy loss; then, they train the student network with the dual goals of prediction of correct label and matching of teacher's probability prediction.
 Surprisingly, despite that the teacher and the student models have
 \textit{identical network structure}, and the \textit{same training data} is used for teacher training and knowledge transfer process, they reported that
 with this BAN approach, the student outperforms the teacher network accuracy consistently in various conventional image classification setups, \eg, DenseNets \cite{huang2017densely} on CIFAR-10 and CIFAR-100 \cite{krizhevsky2009cifar}.
 The student models were found to have {\em better  generalization}. This is attributed to the distillation of {\em dark knowledge}, \ie, 
 teacher's prediction on the wrong outputs, and {\em importance weighting}, \ie, teacher's confidence on the correct outputs.
 Recently, 
 Zhu and Li 
 \cite{allen2020msdistillation}
 presented a rigorous analysis on this improved generalization.
 From the perspective of {\em multi-view} data structure, they argue that the BAN approach can be viewed as a combination of implicit ensemble and knowledge distillation
 \cite{hinton-distill}, enabling the student model to learn multi-view features and eventually achieve better generalization compared to the teacher models which have identical structures.
 Besides the conventional image classification, 
 BAN has been applied in other areas, \eg, multi-task natural language processing \cite{clark-etal-2019-bam}.
 
 \subsection{Motivation and Our Contributions}
 This work is motivated by the empirical results and theoretical analysis presented by
 \cite{furlanello2018bornagain} and \cite{allen2020msdistillation}. Both works suggested BANs can achieve improved generalization without modification to the network structure, which could be extremely useful for existing FSC models, especially, under domain shift.
 In particular, {\bf our first contribution} is to propose BAN episodic training for DG-FSC.
 Note that previous work has focused on applying BAN in conventional supervised training 
 \cite{furlanello2018bornagain, allen2020msdistillation, tian2020rethinkingsd-fsc}, and our work on applying BAN in \textit{episodic training} is novel.
 %in order to handle the DG-FSC task in the meta-learning \cite{finn2017maml} framework.
 %\linebreak
 In Sec.\ref{3.3}, we discuss the subtleties in BAN episodic training,  perform a rigorous study to show that BAN can lead to models with improved generalization on novel tasks sampled from an unseen domain.
 Furthermore, we also validate that BAN enables learning of more compact features with a lower intra-class to inter-class variance ratio which is useful for few-shot learning as discussed in \cite{gold-icml2020-unraveling}
 (sSee Linear Discriminant Analysis (LDA) \cite{mika1999fisher} of features in {Figure \ref{fig1}}).

 Based on the encouraging results investigated
 in Sec. \ref{3.3}, {\bf our second contribution} is to propose Few-Shot BAN (FS-BAN) that addresses the unique issues in DG-FSC.
 Specifically, 
 different from conventional image classification, DG-FSC poses unique challenges that inhibit the improvement of BAN episodic training: 
 (i) Because of limited labeled data in FSC, the teacher model in BAN training may suffer from overfitting, and this degrades the knowledge transferring to the student model; 
 (ii) In DG-FSC, the student model needs to handle unseen domains during the evaluation stage.

%  ; , it is important to imitate such domain shift in the training phase; (iii) The key hyperparameter (\eg, temperature) used in BAN for knowledge transfer is fixed for different seen source domains in training, which would be sub-optimal.
 
 To address the above challenges in BAN for DG-FSC, we propose FS-BAN (Sec. \ref{sec4}) that builds upon the baseline BAN method (Sec. \ref{3.3}). FS-BAN consists of novel multi-task learning objectives:
 % MR
 (i) Mutual Regularization (MR):
 We extend BAN with additional feedback {\em from the student to the teacher}, encouraging the teacher to continue to improve using soft predictions from the student. The student's soft prediction provides additional regularization  to alleviate overfitting in the teacher model. This technique achieves significant improvements in all experiments.
 % MM
 (ii) Mismatched teachers (MM):
 To address domain shift, we propose a technique of {\em mismatched teacher}: a teacher model which is trained on a domain different from that of the current training task.
 Our proposed mismatch teacher is an imitation procedure so that an FSC model has exposure to domain shift during the training stage. We show in experiments that this imitation in training leads to a better generalization of unseen domains and achieves better domain robustness.
 % MCT
 (iii) Meta-control temperature (MCT):
 Temperature is an important parameter to control the distillation of knowledge in BAN training \cite{hinton-distill}.
 It is usually {regarded} as a hyperparameter and manually pre-set to a fixed value for the entire training (regardless of different domains and tasks).
 In contrast, we propose to \textit{meta-learn the temperature} during training to improve adaptation to diverse domains. 
 
 The proposed FS-BAN can be readily applied to  
 existing FSC models 
 without modification of the structure. Experiment results show that FS-BAN achieves new state-of-the-art results for DG-FSC on six benchmark datasets, with three popular FSC baseline models. We further show in comprehensive ablation studies that the different learning objectives in FS-BAN indeed address these challenges proposed above.
 
 {Our contributions in this paper are summarized as}:
 
 \begin{enumerate}
    \item As a pioneer work, we propose BAN episodic training as our first contribution (Sec.\ref{3.3}). 
    We carefully study its effectiveness for DG-FSC and compare it to related work. We empirically validate its improved generalization. 
    
    %and limitations for existing FSC models on unseen domains, which has not been done before.
    
    \item As our second contribution, we propose FS-BAN for DG-FSC (Sec.\ref{sec4}).
    FS-BAN consists of multi-task learning objectives that can better address the unique challenges posed by DG-FSC: few labeled support data in an episode and domain shift in the testing phase.
    FS-BAN overcomes the challenges, and such efforts have not been done before.
    
    \item We conduct extensive experiments and show that FS-BAN consistently improves three baseline FSC models on six public datasets. 
    Our approach outperforms the state-of-the-art in both the conventional FSC and DG-FSC setups. 
    We also perform detailed ablation studies to demonstrate the effectiveness of FS-BAN.
 \end{enumerate}

% related work

\section{Related Works}
 \label{sec2}
 In this section, we perform a literature review from different perspectives, as our work involves FSC, domain generalization, and effective knowledge transfer. We highlight the different and challenging problem setups compared to closely related traditional FSC and domain generalization tasks. 
 
 %We also demonstrate the motivation of our study (as in Sec. \ref{3.3}) and the proposed method (as in Sec. \ref{sec4}). 
 
  %%%%%%%%%%%%%%%%%%%%%%%%%%%%%%%
 \begin{figure*}[t]
    \centering
    \includegraphics[width=0.99\textwidth]{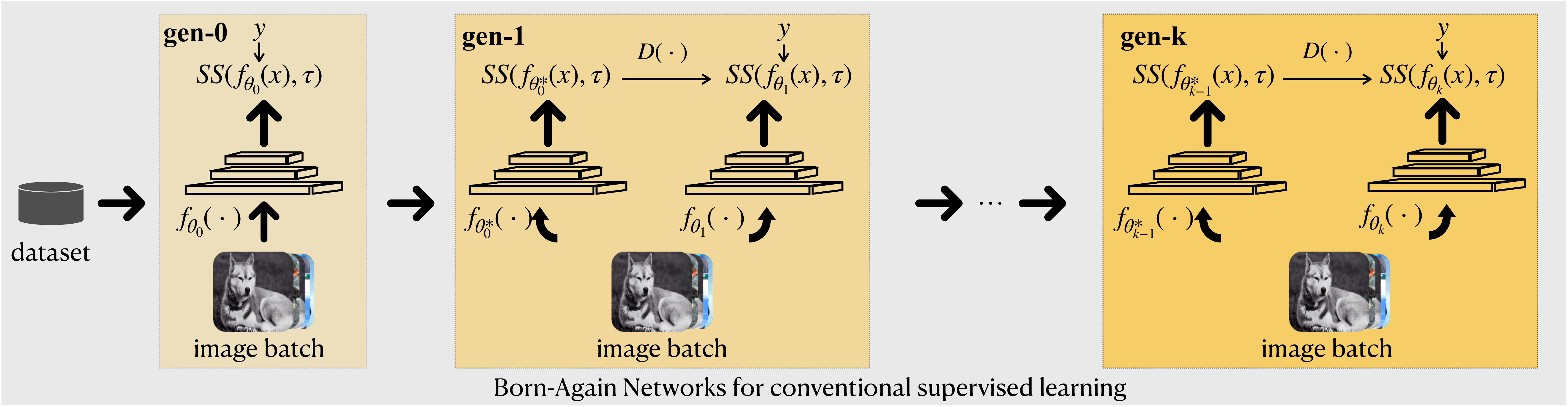}
    \caption{
    {
    % \color{blue}
    In conventional supervised learning, 
    }
    BAN samples a batch of images $\{(x,y)\in (\mathcal{X}, \mathcal{Y})\}$ of all categories in the dataset and distills the knowledge from the teacher model to the student in each generation.
    {
    % \color{blue}
    In related work, Tian \etal \cite{tian2020rethinkingsd-fsc} conducted the born-again process in generations to obtain a powerful backbone network and transferred it to the downstream FSC task.
    }
    \vspace{-4 mm}
    }
    \label{fig_ban_conventional}
 \end{figure*}
 %%%%%%%%%%%%%%%%%%%%%%%%%%%%%%%
 
 \subsection{Metric Learning for Few-Shot Classification} 
 FSC \cite{finn2017maml, snell2017prototypical} models aim to recognize novel classes given few labeled data. Among them, metric learning based methods \cite{snell2017prototypical, vinyals2016matching, sung2018relationnet, oreshkin2018tadam} learn to compare the relation between the unlabeled query data and the labeled support data. The prediction result of each query image is a confidence (probability) distribution assigned to each category that belongs to a training task. Metric learning based ideas have attracted a fair amount of attention on FSC tasks. Meanwhile, there is no need to further fine-tune the model parameters or select the hyperparameters in test time \cite{tian2020rethinkingsd-fsc, chen2019closer}.

 In this paper, we set our experiments to focus on three popular metric-based FSC models as baseline methods, similar to a recent work \cite{tseng2020cross}: MatchingNet \cite{vinyals2016matching}, RelationNet \cite{sung2018relationnet} and Graph Neural Network (GNN) \cite{garcia2017gnn} due to their simplicity and easy implementation.
 However, these models often fail to make predictions on novel tasks from unseen domains, due to the domain shift \cite{chen2019closer,zhang2020deepemd, triantafillou2019meta-dataset} and overfitting on the base classes data from the source domains seen in training. Therefore, our proposed FS-BAN builds up on their models and aims to obtain further improvement and generalization.
 
 \subsection{Domain Generalization FSC} 
 Traditional domain adaptation (DA) problem often enables the model to learn with sufficient unlabeled data from the target domain \cite{li2016adaBN, li2021cvpr21-da-1, yue2021cvpr21-da-2, abdollahzadeh2021revisit, gong2022diffpose} in the training stage. Therefore, the domain discrepancy between the source and the target domains could be explicitly reduced. 
 Different from DA, domain generalization (DG) \cite{li2017dg-problem, pandey2021cvpr21-dg-1} aims to learn good feature representations that generalize well on unseen domains in test time \cite{Zhao_2022_CVPR_fsig, Zhao_2022_NeurIPS_fsig, Zhao_2023_CVPR_fsig}.
 For the traditional supervised classification tasks, \cite{li2019episodic_random, ghiasi2018dropblock} propose to add regularization objectives in the training stage to improve the generalization performance. However, the label space for training and testing is shared therefore there is still prior knowledge of the target domain.
 
 In DG-FSC, models are needed to recognize samples of novel categories from unseen domains, given only few (\eg, 5-shot) labeled support data.
 Very recently, \cite{tseng2020cross} applies the 
 learned feature-wise transformation layer (LFT) \cite{perez2017film} to modulate the channel-wise scale and shift parameters, trying to produce diverse and entangled feature representations of different domains.
 \cite{sun2020explanation} applies the explanation-guided layer-wise relevance propagation (LRP) to enhance the discriminative features during training with multiple seen domains.
 \cite{zhao2021domain-adaptive-fsl} address a similar problem but they use the unlabelled data from target domains in the training phase. 

 Our proposed FS-BAN, differently, aims to improve the generalization of FSC models for episodic training in DG-FSC setup, by less overfitting to hard targets and it is more robust to arbitrary unseen domains, with disjoint label space (\eg, train on Cars domain \cite{fei-fei-li2013-cars} but test on Birds species \cite{hilliard2018cub-split}) during evaluation. Our setup is more challenging compared to conventional supervised learning but closer to the real-world applications and model deployment environment.

 %%%%%%%%%%%%%%%%%%%%%%%%%%%%%%%
 \begin{figure*}[t]
    \centering
    \includegraphics[width=0.99\textwidth]{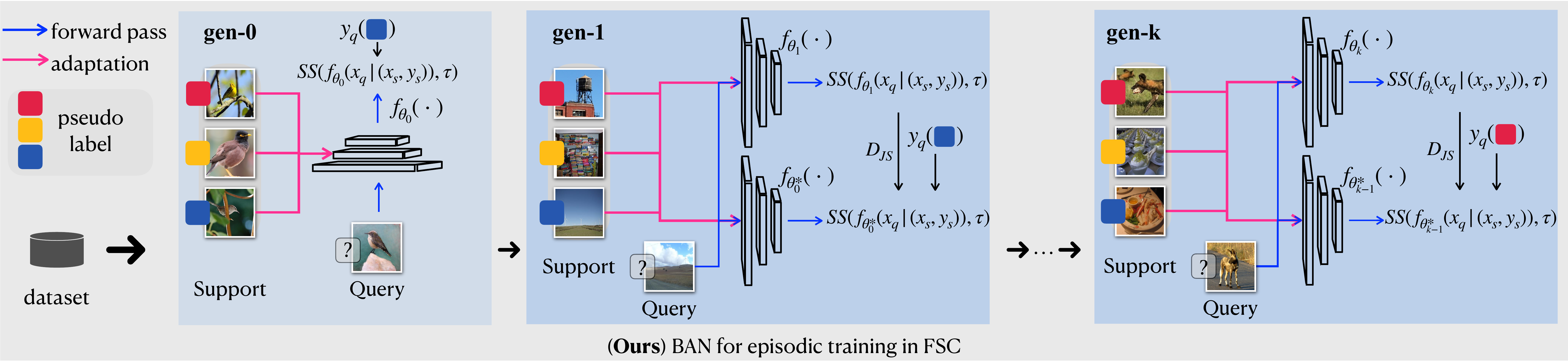}
    \caption{
    Our proposed BAN episodic training. 
    A task \(\mathcal{T}\) with $N_{w}$ 
    categories is sampled
    (here $N_{w}=3$).
    The support set of \(\mathcal{T}\) is applied to 
    adapt the teacher and student models.
    Then, the teacher conditioning on the support set predicts the query samples of \(\mathcal{T}\)
    and transfers the knowledge to the student. 
    {
    % \color{blue}
    Compared to \cite{tian2020rethinkingsd-fsc} (see Sec. \ref{3.3}) that adopts the transfer learning approach, we directly apply BAN in episodic training that simulates the realistic setting in the evaluation phase for FSC.
    }
    }
    \label{fig_ban_episodic}
    \vspace{-4mm}
 \end{figure*}
 %%%%%%%%%%%%%%%%%%%%%%%%%%%%%%%
 
 \subsection{Knowledge Distillation and Born-Again Network} 
 Knowledge distillation (KD) \cite{hinton-distill, zhao2022revisiting} often aims to transfer the ``knowledge'' of a larger and stronger machine learning model (\textit{the teacher}) learned on a large-scale dataset, to another compact model (\textit{the student}) with a small training dataset \cite{lihonglin2023task, Zhao_2023_arxiv_watermark_dm}. 
 KD has shown empirical benefits in some applications, \eg, model compression \cite{wang2019private} and transfer learning \cite{yin2020dreaming}.
 Usually, KD method can train the student network that benefits from the teacher's knowledge and obtains a good performance. 
 
 Born-Again Network (BAN) \cite{furlanello2018bornagain} is a special case of KD that transfers knowledge from well-trained teacher(s) to the student \textit{with an identical network structure and training data}. 
 Taking advantage of this, BAN can generate multiple generations by repeating the knowledge-transfer process (we discuss this Sec. \ref{sec3}). 
 Surprisingly, previous works \cite{furlanello2018bornagain, clark-etal-2019-bam} discover that the student can outperform the teacher consistently in terms of prediction accuracy on conventional supervised learning tasks, which suggests improved generalization to the test data.
 Recently, \cite{tian2020rethinkingsd-fsc} applies BAN in conventional classification task
 (\ie, Figure \ref{fig_ban_conventional}) to obtain a backbone network. Then, they apply the standard transfer learning pipeline on the student model to handle the downstream single-domain FSC tasks. 
 In this work, we design FS-BAN for DG-FSC that takes the advantage of BAN the improved generalization without the need for modification to the network structure and additional training data. Compared to the similar work \cite{tian2020rethinkingsd-fsc}, our designs are clearly different, as shown in Figures \ref{fig_ban_episodic} and Figure \ref{FS-BAN}. Comparison results with \cite{tian2020rethinkingsd-fsc} show the superiority of our designs (see Table
 \ref{table4}).

% section 3

 \section{Preliminary}
 \label{sec3}
 In this section, we discuss the concepts of BAN and DG-FSC. 
 Concretely, in Sec. \ref{3.1}, we review the mechanism of BAN in conventional supervised image classification;
 in Sec. \ref{3.2}, we formulate the DG-FSC problem setup and the episodic training process of existing FSC models.
 
 \subsection{BANs for Conventional Supervised Image Classification}
 \label{3.1}
  We follow the definition of BAN in conventional classification problems \cite{furlanello2018bornagain}. Consider a dataset containing image samples \(\mathcal{X}\) and true labels \(\mathcal{Y}\).
%   of \(\mathcal{C}\) categories,
%   \ie, \(\mathcal{Y} \in \mathbb{R}^{\mathcal{C}}\).
  Generally, the prediction of the input samples \(\mathcal{X}\) is parameterized by a network \(f_{\theta_0}(\mathcal{X})\). \(f_{\theta^*_0}(\cdot)\) is called \textit{the teacher} network and it can be obtained by minimizing the cross-entropy loss to the ground truth labels:
  \begin{equation}
    \label{eq1}
      \theta^*_0 = \arg\min_{\theta_0}\mathcal{L}_{ce}(\mathcal{Y}, \hat{\mathcal{Y}}^{\theta_0}),
  \end{equation}
 where $\hat{\mathcal{Y}}^{\theta_0}=SS(f_{\theta_0}(\mathcal{X}), \tau)$. \(SS(\cdot, \tau)\) is the SoftMax function with a temperature \(\tau\) over $N$ training classes:
 \begin{equation}
    {SS}(z, \tau) =  \frac{e^{z/\tau}}{\sum_{c=1}^{N}e^{z_{c}/\tau}},
    \label{eq_ss}
 \end{equation}
 where we assume $z$ is input to the SoftMax layer. Normally, Eqn. \ref{eq_ss} is considered to soften or harden the soft predictions when \(\tau>1\) or \(\tau<1\). 
 As Figure \ref{fig_ban_conventional}, BAN enables another model (\(f_{\theta_1}(\cdot)\), \textit{the student}) to exploit the rich information contained in the predicted probability distribution of the teacher, by minimizing the distance ($D$)
 between the output distribution of the teacher \(f_{\theta_0^*}(\cdot)\) and that of the student \(f_{\theta_1}(\cdot)\):
  \begin{equation}
     \mathcal{L}_{ce}(\mathcal{Y}, \hat{\mathcal{Y}}^{\theta_1}) + D(SS(f_{\theta_1}(\mathcal{X}), \tau), SS(f_{\theta_0^*}(\mathcal{X}), \tau)),
  \end{equation} 
  where the first term is the classification loss to the one-hot ground truth, and the second term employs the soft prediction of the fixed teacher model for knowledge transfer.
  
  Since the student has the identical structure and training data of the teacher, this \textit{born-again} process can be applied sequentially with multiple generations: In \(k\)-th generation (gen-\(k\), $k>1$), the student \(f_{\theta_{k}}(\cdot)\) is trained to optimize a sum of cross-entropy loss and the distance between its prediction and the soft targets from the student obtained in gen-($k$-$1$):
  \begin{equation}
    \label{eq3}
    \mathcal{L}_{ce}(\mathcal{Y}, \hat{\mathcal{Y}}^{\theta_k}) + D(SS(f_{\theta_{k}}(\mathcal{X}), \tau), SS(f_{\theta_{k-1}^*}(\mathcal{X}), \tau)).
  \end{equation} 
  The student $f_{\theta_{k-1}^*}(\cdot)$ obtained in ($k$-$1$)-th generation now becomes the new teacher. In particular, \(f_{\theta^*_{0}}(\cdot)\) indicates the first teacher that is trained with only cross-entropy loss to one-hot labels in gen-0.
  Interestingly, previous work \cite{furlanello2018bornagain, tian2020rethinkingsd-fsc} reported {\em improved generalization} of student network with BAN training in this conventional supervised learning setup, which motivates us to investigate BAN for DG-FSC. We discuss it in Sec. \ref{3.3}.

 \subsection{Metric-based Models for DG-FSC}
 \label{3.2}
 Here, we discuss the problem setup in this work.
 We follow the popular meta-learning algorithms \cite{finn2017maml, snell2017prototypical, sung2018relationnet,  guo2020awgim, vinyals2016matching, garcia2017gnn,  rusu2018-leo} to define episodic training for DG-FSC. 
 
%  (\ie, )
 \textbf{Episodic training.}
 We denote the input images as \(\mathcal{X}\) and the corresponding labels as \(\mathcal{Y}\).
 In each training iteration, instead of sampling a batch of images with their true labels directly (as in conventional supervised learning), we sample an
 \(N_{w}\)-Way (number of classes) \(N_{s}\)-Shot (number of labeled samples per class) task \(\mathcal{T}\) of a source domain \(\mathcal{D}\) from several seen domains \(\{\mathcal{D}_{1}, \mathcal{D}_{2}, \dots, \mathcal{D}_{n}\}\) \cite{tseng2020cross}.
 Each \(\mathcal{T}\) consists of a support set \(\mathcal{S}=\{(\mathcal{X}_s, \mathcal{Y}_s)\}\), and a query set \(\mathcal{Q}=\{(\mathcal{X}_q, \mathcal{Y}_q)\}\). 
 The support set \(\mathcal{S}\) and the query set \(\mathcal{Q}\) are formed by randomly selecting \(N_{s}\) and \(N_{q}\) samples of each of \(N_{w}\) categories (usually, \(N_{w}=5\)), respectively. In this context, the batch size is one task (or an episode), and the samples in \(\mathcal{S}\) and \(\mathcal{Q}\) are pseudo-labeled which will change in different episodes. 

 \textbf{Metric learning based FSC.}
 Suppose a metric-based FSC model \(f\) is parameterized by \(\theta\). For each sampled task \(\mathcal{T}\), \(f_{\theta}(\cdot)\) firstly extracts the feature embeddings of both support \(\mathcal{S}\) and query samples \(\mathcal{Q}\), then it predicts the label of each query sample by comparing its relation to support sample features (\ie, conditioned on the labeled support set):
 \begin{equation}
    \label{eq4}
    \hat{\mathcal{Y}}_q^{\theta}= SS(f_{\theta}(\mathcal{X}_q | (\mathcal{X}_s, \mathcal{Y}_s)), \tau),
 \end{equation}
 where \(\hat{\mathcal{Y}}_q^{\theta}\) is the prediction results of query samples over $N_{w}$ classes. Generally, we aim to minimize the prediction error on the query set with cross-entropy loss w.r.t. one-hot labels:
 \begin{equation}
     \mathcal{L}_{ce}(\mathcal{Y}_q, \hat{\mathcal{Y}}_q^{\theta}).
     \label{eq5}
 \end{equation}
 In the testing phase, we evaluate the accuracy of the query set of tasks sampled from \textit{novel} classes of \textit{unseen} domains. 
 We follow the DG setup \cite{tseng2020cross, sun2020explanation, li2019episodic_random} that we do not approach any samples from unseen domains in the training phase.
 Therefore, our FSC models are expected to learn robust and discriminative knowledge that can be well transferred to other domains. 
 Note that in this DG-FSC setup, the label spaces of source domains and target unseen domains are disjoint, different from some recent DG literature \cite{li2018rotate-domain, li2018mmd-aae, li2019episodic_random}.

 %%%%%%%%%%%%%%%%%%%%%%%%%%%%%%%
 % Table 1
 \begin{table}[t]
    \centering
    \caption{
    Accuracy (\%) of the proposed BAN episodic training (Figure \ref{fig_ban_episodic}) for DG-FSC.
    Model is trained with 5-Way 1-Shot tasks of base classes of miniImageNet. RelationNet \cite{sung2018relationnet} is the baseline model and ResNet-10 \cite{he2016resnet} is the backbone network. 
    \textbf{Top}: Model tested on novel classes of different unseen domains.
    \textbf{Bottom}: Performance of BAN episodic training in different generations on novel classes of miniImageNet (seen) and CUB (unseen).
    See Sec. \ref{3.3}. for more details.}
    \begin{adjustbox}{width=\columnwidth,center}
    \begin{tabular}{l c|c|c|c|c}
        \toprule
         % \multirow{2}{4em}{Method} & 
         % {mini $\mapsto$} & {mini $\mapsto$} & {mini $\mapsto$} & {mini $\mapsto$} \\
         % & {CUB} & {Cars} & {Places} & {Plantae}\\
         \textbf{Method} & \textbf{Source}
         & {CUB} & {Cars} & {Places} & {Plantae}\\
         \hline
         RelationNet (gen-0) & miniImageNet & $42.44$ & $29.11$ & $48.64$ & $33.17$  \\
         +BAN (gen-1) & miniImageNet & $\bm{43.35}$ & $\bm{29.71}$ & $\bm{51.30}$ & $\bm{33.81}$ \\
        \bottomrule
    % \end{tabular}
    % % \end{adjustbox}
    % % \begin{adjustbox}{width=0.99\columnwidth,center}
    % \begin{tabular}{l|c|c|c|c|c}
    \toprule
        \textbf{Dataset} & gen-0 & gen-1 & gen-2 & gen-3 & gen-4 \\
        \hline
    miniImageNet (base $\mapsto$ novel) & $57.80$ & $60.45 $ & $60.79$ & \bm{$61.47$} & $61.39$\\
    % Improvement percentage (\%) & - & \bm{$4.58$} & $0.56$ & $1.11$ & $-0.13$ \\\hline
    miniImageNet $\mapsto$ CUB & $42.44$ & $43.35$ & $43.64$ & \bm{$43.92$} & $43.87$\\
    % Improvement percentage (\%) & - & \bm{$2.14$} & $0.67$ & $0.64$ & $-0.11$  \\
    \bottomrule
    \end{tabular}
    \end{adjustbox}
    \vspace{-3 mm}
    \label{table1}
 \end{table}
 %%%%%%%%%%%%%%%%%%%%%%%%%%%%%%%
 %%%%%%%%%%%%%%%%%%%%%%%%%%
 \begin{figure}[ht]
     \centering
     \includegraphics[width=0.241\textwidth]{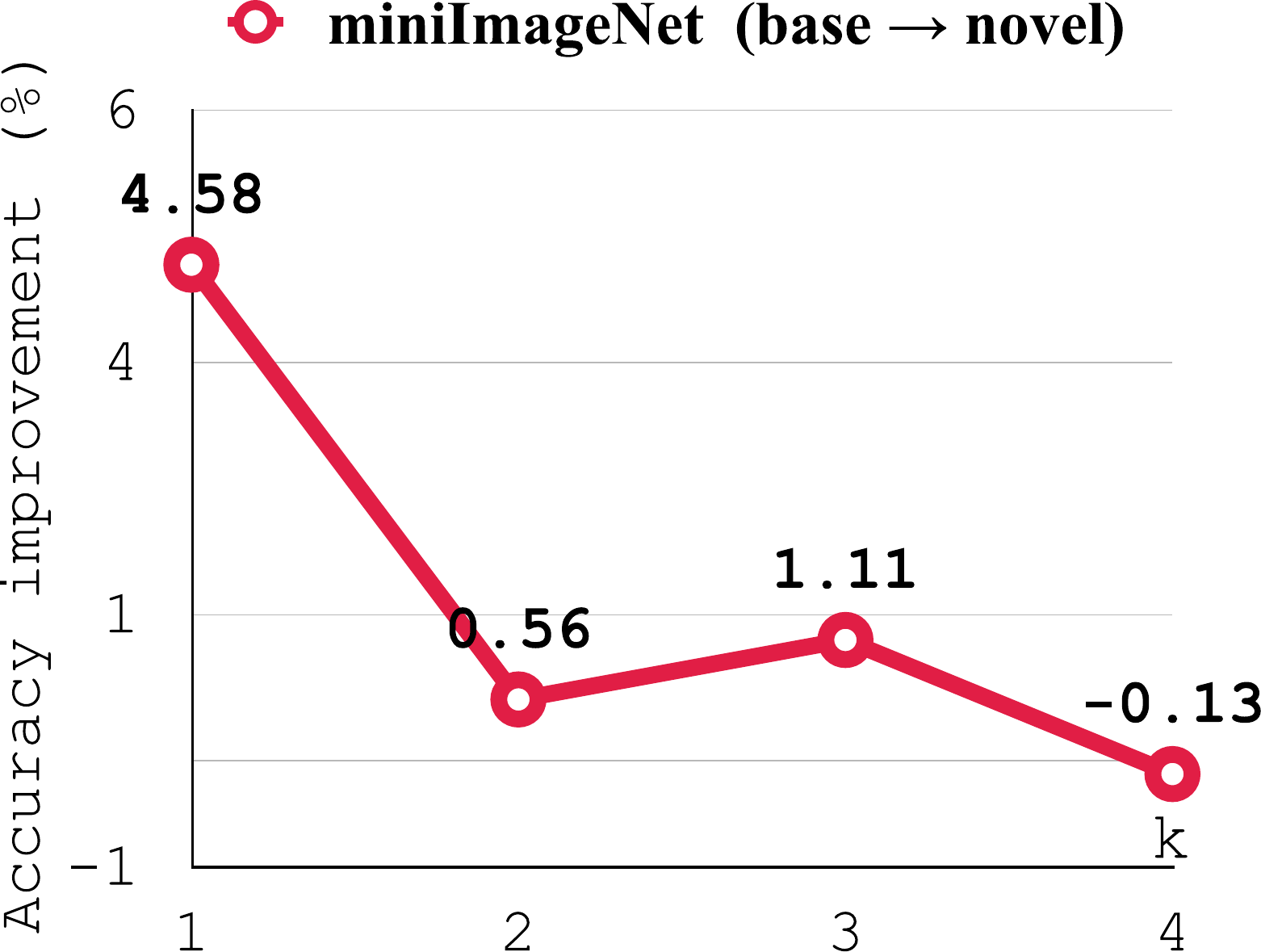}
     \includegraphics[width=0.241\textwidth]{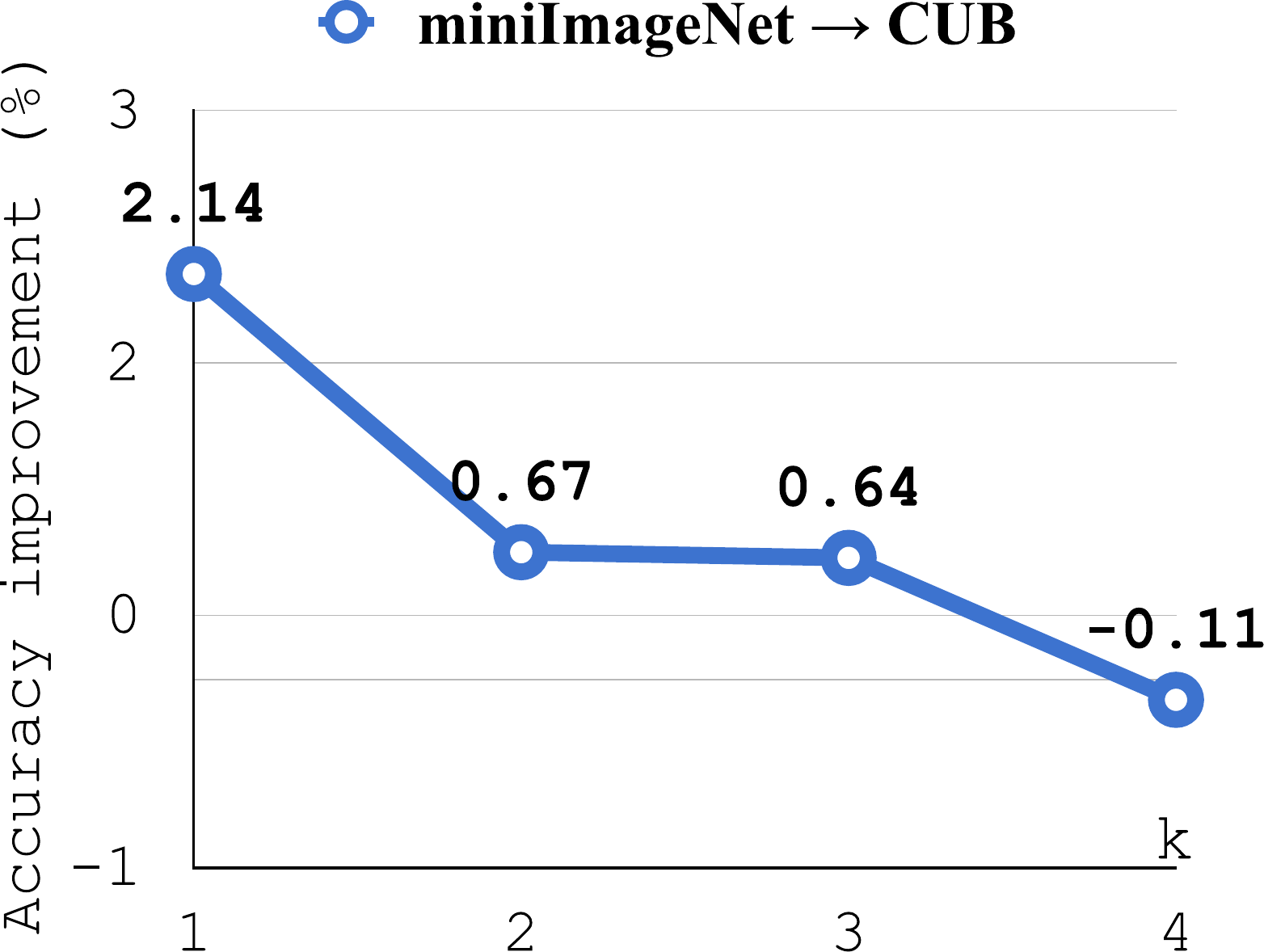}
     \caption{Accuracy improvement (\%) of gen-(k-1) $\mapsto$ gen-k of Table \ref{table1}. It is clear that the major gain is obtained at gen-0 $\mapsto$ gen-1. The deeper generations come with expensive training costs and lead to diminishing increment, and the negative impact is observed after the empirical optimal generation (gen-3 in Table \ref{table1}). We note that this observation is consistent with that of BAN in conventional supervised learning \cite{furlanello2018bornagain, tian2020rethinkingsd-fsc}.}
     \label{ban_percent}
     \vspace{-2 mm}
 \end{figure}
 %%%%%%%%%%%%%%%%%%%%%%%%%%%%%%%

% section 4 ban episodic training
 \section{Born-Again Episodic Training for DG-FSC}
 \label{3.3}
 \textbf{BAN episodic training.} Motivated by the theoretical analysis in \cite{allen2020msdistillation}, and improved generalization observed in conventional supervised learning \cite{furlanello2018bornagain}, in this section, we propose BAN episodic training for DG-FSC. We conduct a rigorous study and show the effectiveness of BAN for the existing FSC model under domain shift, which motivates us to propose FS-BAN (discussed in the next section). 
 
 As Figure \ref{fig_ban_episodic} and description in Sec. \ref{3.2}, in each training iteration of DG-FSC during the \(k\)-th generation of BAN, rather than sampling a batch of images of all classes, we instead sample a task \(\mathcal{T}\) with  
 $N_{w}$ categories. 
 We apply the support set of \(\mathcal{T}\) to adapt both the teacher and student models.
    Then,  
    the models conditioning on the support set are used to predict query samples of \(\mathcal{T}\).
    After that,  
 %For both the teacher and the student, we firstly extract the features of support set and the query set. Then, for each query sample, we obtain the predicted probability distribution by comparing its relation to the support set samples. Finally, 
 similar to Eqn. \ref{eq3}, we optimize the student network \(f_{\theta_{k}}(\cdot)\) by leveraging the one-hot label and the soft targets predicted by the teacher network \(f_{\theta_{k-1}^*}(\cdot)\) on the same query set \(\mathcal{Q}\):
 \begin{equation}
    \label{eq6}
    \mathcal{L}_{BAN} = \lambda_1\mathcal{L}_{ce}(\mathcal{Y}_q, \hat{\mathcal{Y}}_q^{\theta_k}) +  \lambda_2\tau^2\DJS(\hat{\mathcal{Y}}_q^{\theta_k}, \hat{\mathcal{Y}}_q^{\theta_{k-1}^*}),\\
 \end{equation}
 where \(\lambda_1\) and \(\lambda_2\) are coefficients of the weighted sum, and \(\hat{\mathcal{Y}}_q^{\theta_{k}}=SS(f_{\theta_{k}}(\mathcal{X}_q | (\mathcal{X}_s, \mathcal{Y}_s)), \tau)\). 
 We use \textit{JS} divergence \cite{fuglede2004jsd} as the distance metric. Meanwhile, since the magnitudes of the gradients produced by the soft targets are scaled by \(\frac{1}{\tau^2}\), we multiply the second term of Eqn. \ref{eq6} by \(\tau^2\) to maintain the balance \cite{hinton-distill}.
 In the meta-testing phase, the temperature is set to \(\tau\)=1 to evaluate the accuracy of novel tasks. The teacher is discarded, hence the outcome of BAN episodic training is the student model without any additional parameters.
 \begin{figure*}[ht]
    \centering
    \includegraphics[width=0.99\textwidth]{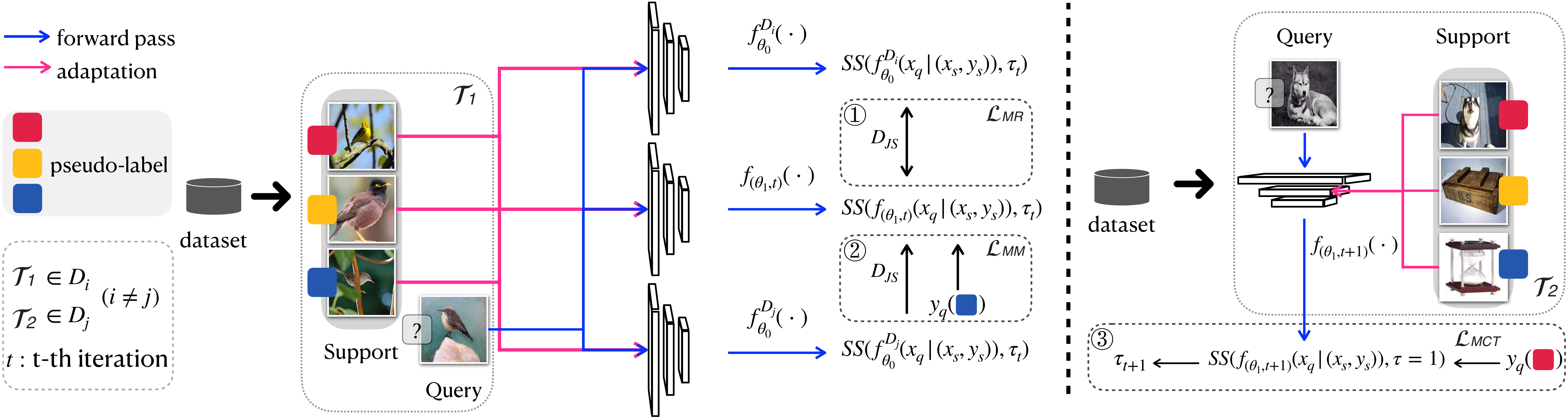}
    \caption{
    Overview of our proposed FS-BAN for DG-FSC.
    \textcircled{\raisebox{-0.9pt}{1}}\textbf{Mutual regularization}. To overcome the potential overfitting of teachers due to limited labeled data in an episode, we propose to regularize the teacher to match the soft distribution from the student.
    \textcircled{\raisebox{-0.9pt}{2}} \textbf{Mismatched teacher}. To explicitly consider domain-shift in training, for a task sampled from domain \(\mathcal{D}_i\), 
    we propose to select a mismatched teacher trained on \(\mathcal{D}_j\) for the knowledge transfer, where \(i\neq j\).
    \textcircled{\raisebox{-0.9pt}{3}} \textbf{Meta-control the temperature}. The temperature \(\tau\) is meta-updated in different iterations by evaluating the performance of the updated student on task from \(\mathcal{D}_j (i \neq j)\). 
    }
    \label{FS-BAN}
    \vspace{-4 mm}
 \end{figure*}  
 %%%%%%%%%%%%%%%%%%%%%%%%%%%%%%%
 
 {\bf Experiment setups.} 
 To validate the effectiveness of the proposed BAN episodic training for DG-FSC, we design two experiment setups: (a) We train the student with one generation, \ie, \(k\)=1, and test its performance on tasks of novel classes from various unseen domains. (b) We evaluate the performance of different born-again generations on novel classes of both seen and unseen domains.
%  {\bf Implementation details.}
 We employ a popular metric-based FSC model RelationNet \cite{sung2018relationnet} as the baseline method in this experiment. Follow \cite{tseng2020cross}, we use ResNet-10 \cite{he2016resnet} as the backbone network. We train each student network with 800 epochs (100 tasks in each epoch) of 5 generations. To enable the episodic training, in each iteration, we sample a 5-Way 1-Shot task from the base classes of miniImageNet \cite{ravi2016optimization}. In the testing stage, we randomly sample 1000 tasks from novel classes of either miniImageNet or different unseen domains to evaluate the performance of BAN in setup (a) and setup (b), with the average accuracy reported. We include the detailed dataset information in Sec. \ref{sec5}.
 %%%%%%%%%%%%%%%%%%%%%%%%%%%%%%%
 % Table 1
 \begin{table}[t]
    \centering
    \caption{
    {
    % \color{blue}
    Accuracy (\%) of BAN via transfer learning \cite{tian2020rethinkingsd-fsc} and the proposed BAN episodic training for DG-FSC. 
    For both methods, ResNet-10 \cite{he2016resnet} is the backbone network and ProtoNet \cite{snell2017prototypical} is used as the classifier head for a fair comparison. 
    \textbf{Top:} miniImageNet (base $\mapsto$ novel),
    \textbf{Bottom:} miniImageNet $\mapsto$ CUB (unseen).
    Other experiment setups are the same as Table \ref{table1}. See detailed analysis in Sec. \ref{3.3}.
    }
    }
    {
    \begin{adjustbox}{width=0.96\columnwidth,center}
    {
    % \color{blue}
    \begin{tabular}{l|c|c|c|c|c}
    \toprule
        \textbf{Method} & gen-0 & gen-1 & gen-2 & gen-3 & gen-4 \\
        \hline
    BAN transfer learning \cite{tian2020rethinkingsd-fsc} & $42.62$ & $46.61$ & $47.53$ & \bm{$47.92$} & $47.81$\\
    BAN episodic training (Ours) & $50.39$ & $53.10$ & $54.61$ & \bm{$55.08$} & $54.91$\\
    \bottomrule
    % \end{tabular}
    % \begin{tabular}{l|c|c|c|c|c}
    \toprule
        \textbf{Method} & gen-0 & gen-1 & gen-2 & gen-3 & gen-4 \\
        \hline
    BAN transfer learning \cite{tian2020rethinkingsd-fsc} & $38.19$ & $39.07$ & \bm{$40.78$} & $40.66$ & $40.32$\\
    BAN episodic training (Ours) & $38.26$ & $39.66$ & $40.63$ & \bm{$41.10$} & $40.77$\\
    \bottomrule
    \end{tabular}}
    \end{adjustbox}
    \label{table1c}
    }
    \label{table-apple-to-apple}
    \vspace{-6 mm}
 \end{table}
 %%%%%%%%%%%%%%%%%%%%%%%%%%%%%%%
 
 {\bf Results and analysis.}
 The experiment results are shown in Figure \ref{fig1} (qualitatively), Table \ref{table1}, and Figure \ref{ban_percent} (quantitatively). Empirically, our observations can be summarized as follows:
 \begin{enumerate}
     \item {Table \ref{table1} \textbf{(Top)}}: Similar to the observation in conventional supervised learning, BAN episodic training can achieve consistent improvement on various novel unseen classes and unseen domains. This suggests the potential of BAN in boosting the generalization of other FSC models in domain generalization setups. 
     \item {Table \ref{table1} \textbf{(Bottom)}}: Multiple BAN generations lead to diminishing improvements. Compared to the born-again learning process of gen-0 $\mapsto$ gen-1, the improvement becomes small in deeper generations. We even observe the performance drop after the empirical optimal generation. Similar observations are also found in other applications of conventional supervised learning \cite{tian2020rethinkingsd-fsc, yang2019aaai-second}. See detailed analysis in Figure \ref{ban_percent}.
     \item Visualization: We extract and analyze the features by the backbone network of a novel task during evaluation. Compared to the baseline model, we observe that BAN can lead to more discriminative features with better decision boundaries, see details in Figure \ref{fig1}. 
 \end{enumerate}

 % %
 % Interestingly, all these empirical improvements and limitations observed in our experiments are in line with that of conventional supervised learning, which motivates us to explore additional improvements of BAN for DG-FSC.
 % %
 
 % In the next, we aim to improve the baseline BAN for DG-FSC from different perspectives. In order to pursue an efficient learning process, we exploit the major gain at gen-0 $\mapsto$ gen-1 and only train one generation student (\ie, $k$=$1$ in Eqn. \ref{eq6}) in the rest part of the paper.
 {
 % \color{blue}
 \textbf{Comparison with BAN transfer learning for FSC.}
 Recently, Tian \etal \cite{tian2020rethinkingsd-fsc} proposed to adopt BAN training in conventional supervised learning (as Figure \ref{fig_ban_conventional}) to obtain a powerful backbone network as the feature encoder. 
 Then, in evaluation, they transfer it to the unseen FSC task ($\mathcal{T}$), extract features of the support set of $\mathcal{T}$, fit a new classifier, and predict query samples. 
 In Table \ref{table-apple-to-apple}, we compare the proposed BAN episodic training with \cite{tian2020rethinkingsd-fsc}. 
 For a fair comparison, for both methods, ResNet-10 \cite{he2016resnet} is the feature encoder, and ProtoNet \cite{snell2017prototypical} is the classifier, which computes the feature distance between the query and the center of support samples of each class (\ie, the ``prototype'') for prediction.
 We show that our proposed BAN episodic training can achieve competitive performance as \cite{tian2020rethinkingsd-fsc} in different (DG-)FSC setups.
 On the other hand, episodic training attempts to simulate a realistic setting in evaluation by learning to solve FSC tasks, and it has been shown very useful to tackle novel, unseen classes given limited labeled data \cite{snell2017prototypical, sung2018relationnet, vinyals2016matching, garcia2017gnn}. 
 Therefore, 
 we are motivated to apply BAN directly in episodic training for (DG-)FSC, as Figure \ref{fig_ban_episodic}. 
 \textbf{Critically}, in contrast to Tian \etal \cite{tian2020rethinkingsd-fsc}, taking advantage of episodic training, we do not modify the network structure or remove/add any layers during the entire training/test phase, and the classifier of our proposed method is compatible with many existing FSC models, which potentially can achieve better performance (see experiments in Sec. \ref{sec5}).
 }

 Next, we propose our improved method of BAN episodic training to tackle unique tasks in DG-FSC.
 In order to pursue an efficient learning process and prevent the computationally expensive sequential training, we exploit the major gain of BAN episodic training at gen-0 $\mapsto$ gen-1 and only train one generation student (\ie, $k$=$1$ in Eqn. \ref{eq6}) in the rest of the paper.

% section 5 method

 %%%%%%%%%%%%%%%%%%%%%%%%%%%%
 % Table 2
\setlength{\tabcolsep}{1 mm}
\renewcommand{\arraystretch}{1}
\begin{table*}[t]
    \scriptsize
    \centering
    \caption{Meta-test accuracy (\%) of DG-FSC with our proposed FS-BAN (Figure \ref{FS-BAN}). 
    %man_sept
    We follow the experiment setup as in
    \cite{tseng2020cross}.
    We let All=\{miniImageNet, CUB, Cars, Places, Plantae\} be the union of all domains for training and testing.
    In \textit{training phase}, we sample tasks from multiple seen domains, \eg, All$\backslash$ \{CUB\}.
    In \textit{testing phase}, we evaluate the model on tasks sampled from the leave-one-out selected unseen domain, \eg, CUB.
    miniImageNet is always the source domain. FS-BAN-lite indicates that we do not include $\mathcal{L}_{MCT}$ in FS-BAN since it requires more GPU memory for training.}
    \begin{adjustbox}{width=0.99\textwidth,center}
    \begin{tabular}{l c c c c c c c c}
         \toprule
            \multirow{2}{4em}{{Method}}
            & \multicolumn{2}{c}{{All$\backslash$ \{CUB\}$\mapsto$CUB}} 
            & \multicolumn{2}{c}{{All$\backslash$ \{Cars\}$\mapsto$Cars}} 
            & \multicolumn{2}{c}{{All$\backslash$ \{Places\}$\mapsto$Places}} 
            & \multicolumn{2}{c}{{All$\backslash$ \{Plantae\}$\mapsto$Plantae}} \\ \cline{2-9}
            & {{5-Way 1-Shot}} & {{5-Way 5-Shot}} & {{5-Way 1-Shot}} & {{5-Way 5-Shot}} & {{5-Way 1-Shot}} & {{5-Way 5-Shot}} & {{5-Way 1-Shot}} & {{5-Way 5-Shot}}\\
            \hline
        
        MatchingNet \cite{vinyals2016matching}
        & \(37.90 \pm 0.55\) & \(51.92 \pm 0.80\)
        & \(28.96 \pm 0.45\) & \(39.87 \pm 0.51\)
        & \(49.01 \pm 0.65\) & \(61.82 \pm 0.57\)
        & \(33.21 \pm 0.51\) & \(47.29 \pm 0.51\) \\
        
        +FT \cite{tseng2020cross}
        & \(41.74 \pm 0.59\) & \(56.29 \pm 0.80\)
        & \(28.30 \pm 0.44\) & \(39.58 \pm 0.54\)
        & \(48.77 \pm 0.65\) & \(62.32 \pm 0.58\)
        & \(32.15 \pm 0.50\) & \(46.48 \pm 0.52\) \\
        
        +LFT \cite{tseng2020cross}
        & \(43.29 \pm 0.59\) & {$61.41 \pm 0.57$}
        & \(30.62 \pm 0.48\) & {$43.08 \pm 0.55$}
        & \(52.51 \pm 0.67\) & \(64.99 \pm 0.59\)
        & \(35.12 \pm 0.54\) & \(48.32 \pm 0.57\) \\
        
        % +MR (Our)
        % & {$45.13 \pm 0.61$} & {$59.75 \pm 0.56$}
        % & {$31.18 \pm 0.55$} & {$42.03 \pm 0.55$}
        % & {$52.97 \pm 0.68$} & {$69.34 \pm 0.57$}
        % & {$39.26 \pm 0.60$} & {$54.61 \pm 0.58$} \\
        
        % +MR+MCT (Our)
        % & \bm{$45.46 \pm 0.59$} & {$61.64 \pm 0.59$}
        % & {$31.67 \pm 0.53$} & {$42.18 \pm 0.56$}
        % & {$53.23 \pm 0.71$} & {$69.80 \pm 0.58$}
        % & {$39.64 \pm 0.64$} & \bm{$56.38 \pm 0.60$} \\
        
        +FS-BAN-lite (Our)
        & {$45.22 \pm 0.65$} & \bm{$62.83 \pm 0.59$}
        & \bm{$31.90 \pm 0.50$} & {$42.44 \pm 0.56$}
        & \bm{$53.53 \pm 0.68$} & {$69.84 \pm 0.55$}
        & {$39.83 \pm 0.62$} & \bm{$54.87 \pm 0.57$} \\

        +FS-BAN (Our)
        & \bm{$45.27 \pm 0.57$} & {$61.34 \pm 0.52$}
        & {$31.71 \pm 0.62$} & \bm{$45.01 \pm 0.57$}
        & {$53.33 \pm 0.67$} & \bm{$70.09 \pm 0.60$}
        & \bm{$40.02 \pm 0.70$} & {$53.89 \pm 0.64$}\\
        \hline
        
        RelationNet \cite{sung2018relationnet}
        & \(44.33 \pm 0.59\) & \(62.13 \pm 0.74\)
        & \(29.53 \pm 0.45\) & \(40.64 \pm 0.54\)
        & \(47.76 \pm 0.63\) & \(64.34 \pm 0.57\)
        & \(33.76 \pm 0.52\) & \(46.29 \pm 0.56\) \\
        
        +FT \cite{tseng2020cross}
        & \(44.87 \pm 0.44\) & \(61.87 \pm 0.39\)
        & \(30.09 \pm 0.36\) & \(40.52 \pm 0.40\)
        & \(48.12 \pm 0.45\) & \(64.92 \pm 0.40\)
        & \(35.53 \pm 0.39\) & \(48.54 \pm 0.38\) \\
        
        +LFT \cite{tseng2020cross}
        & {$48.38 \pm 0.63$} & \(64.99 \pm 0.54\)
        & \(32.21 \pm 0.51\) & \(43.44 \pm 0.59\)
        & \(50.74 \pm 0.66\) & \(67.35 \pm 0.54\)
        & \(35.00 \pm 0.52\) & \(50.39 \pm 0.52\) \\
        
        +LRP \cite{sun2020explanation}
        & \(45.64 \pm 0.42\) & \(62.71 \pm 0.39\)
        & \(30.00 \pm 0.32\) & \(41.05 \pm 0.37\)
        & \(48.74 \pm 0.45\) & \(66.08 \pm 0.40\)
        & \(36.04 \pm 0.38\) & \(48.78 \pm 0.37\) \\
        
        % +MR (Our)
        % & \(47.28 \pm 0.64\) & {$65.01 \pm 0.58$}
        % & {$33.22 \pm 0.57$} & {$43.15 \pm 0.58$}
        % & {$53.00 \pm 0.69$} & {$69.24 \pm 0.56$}
        % & {$38.45 \pm 0.63$} & {$52.74 \pm 0.59$} \\
        
        % +MR+MCT (Our)
        % & \(47.02 \pm 0.58\) & \bm{$65.58 \pm 0.57$}
        % & {$33.40 \pm 0.57$} & {$43.79 \pm 0.58$}
        % & \bm{$54.48 \pm 0.71$} & {$69.71 \pm 0.56$}
        % & {$38.55 \pm 0.63$} & {$52.83 \pm 0.58$} \\
        
        +FS-BAN-lite (Our)
        & \bm{$48.69 \pm 0.65$} & {$65.37 \pm 0.58$}
        & {$33.33 \pm 0.57$} & {$44.35 \pm 0.59$}
        & {$53.43 \pm 0.66$} & \bm{$70.64 \pm 0.56$}
        & {$38.29 \pm 0.62$} & {$53.40 \pm 0.58$} \\

        +FS-BAN (Our)
        & {$47.67 \pm 0.59$} & \bm{$65.55 \pm 0.56$}
        & \bm{$33.43 \pm 0.57$} & \bm{$45.78 \pm 0.57$}
        & \bm{$53.50 \pm 0.68$} & {$69.72 \pm 0.58$}
        & \bm{$38.75 \pm 0.61$} & \bm{$53.55 \pm 0.57$} \\
        \hline
        
        GNN \cite{garcia2017gnn}
        & \(49.46 \pm 0.73\) & \(69.26 \pm 0.68\)
        & \(32.95 \pm 0.56\) & \(48.91 \pm 0.67\)
        & \(51.39 \pm 0.80\) & \(72.59 \pm 0.67\)
        & \(37.15 \pm 0.60\) & \(58.36 \pm 0.68\) \\
        
        +{FT} \cite{tseng2020cross}
        & \(48.24 \pm 0.75\) & \(70.37 \pm 0.68\)
        & \(33.26 \pm 0.56\) & \(47.68 \pm 0.63\)
        & \(54.81 \pm 0.81\) & \(74.48 \pm 0.70\)
        & {$37.54 \pm 0.62$} & \(57.85 \pm 0.68\) \\
        
        +{LFT} \cite{tseng2020cross}
        & \(51.51 \pm 0.80\) & {$73.11 \pm 0.68$}
        & \(34.12 \pm 0.63\) & {$49.88 \pm 0.67$}
        & \(56.31 \pm 0.80\) & \(77.05 \pm 0.65\)
        & {$42.09 \pm 0.68$} & \(58.84 \pm 0.66\) \\
    
        % +MR (Our)
        % & {$51.01 \pm 0.76$} & {$71.61 \pm 0.68$}
        % & {$34.18 \pm 0.65$} & {$48.59 \pm 0.68$}
        % & {$58.57 \pm 0.84$} & {$77.46 \pm 0.62$}
        % & {$40.86 \pm 0.68$} & {$60.06 \pm 0.71$} \\
        
        % +MR+MCT (Our)
        % & \bm{$52.45 \pm 0.82$} & {$72.88 \pm 0.65$}
        % & \bm{$34.51 \pm 0.60$} & {$49.17 \pm 0.67$}
        % & \bm{$58.87 \pm 0.83$} & \bm{$77.98 \pm 0.61$}
        % & {$41.64 \pm 0.73$} & {$60.52 \pm 0.66$} \\
        
        +FS-BAN-lite (Our)
        & \bm{$52.33 \pm 0.77$} & {$72.16 \pm 0.67$}
        & {$34.50 \pm 0.66$} & {$49.29 \pm 0.68$}
        & {$58.86 \pm 0.85$} & {$77.74 \pm 0.62$}
        & {$41.28 \pm 0.68$} & {$61.32 \pm 0.67$} \\

        +FS-BAN (Our)
        & {$52.07 \pm 0.74$} & \bm{$73.70 \pm 0.66$}
        & \bm{$34.87 \pm 0.66$} & \bm{$50.66 \pm 0.65$}
        & \bm{$58.91 \pm 0.87$} & \bm{$78.57 \pm 0.67$}
        & \bm{$41.72 \pm 0.76$} & \bm{$61.85 \pm 0.66$} \\
    \bottomrule
    \end{tabular}
    \end{adjustbox}
    \vspace{-4 mm}
    \label{table2}
\end{table*}
 %%%%%%%%%%%%%%%%%%%%%%%%%%%%
 
 \section{Few-Shot BAN}
 \label{sec4}
 We show in Sec. \ref{3.3} the promising results of BAN episodic training for DG-FSC, which indicate better generalization on novel class tasks from unseen domains during evaluation. 
 However, the improvement of baseline BAN could have been inhibited due to several unique challenges of DG-FSC:
 \begin{enumerate}
     \item The particularity of BAN lies in that the teacher network is trained with an identical structure and the same training data as the student. In this few-shot scenario,  overfitting of the teacher network could degrade the knowledge transferred to the student. 
     \item DG-FSC requires the FSC model to recognize novel tasks from unseen domains that are not accessible during training. Inspired by a recent DG work \cite{li2019episodic_random} for conventional image classification, it is useful to imitate such domain shift during training so that domain robustness can be improved.
     \item The key hyper-parameter temperature $\tau$ in BAN is often pre-set to be a fixed value for different source domains, which could be sub-optimal. For DG-FSC tasks, we expect to find a proper temperature that is suitable for various seen domains and such that the student model can be better generalized to unseen domains.
 \end{enumerate}

 To address the issues, we propose few-shot born-again networks (FS-BAN), including novel multi-task learning objectives with different teacher-student interactions, as Figure \ref{FS-BAN}. We show in experiments that, these challenges are greatly mitigated with a marginal increment of the training cost.

 \subsection{Mutual Regularization}
 \label{4.1}
 We show in Table \ref{table1} that BAN improves DG-FSC before the optimal generation. We attribute this to that the teacher at gen-$k$ ($k>1$) learned the cross-category knowledge \cite{yang2019aaai-second} in the last generation. However, since we train \(f_{\theta_{k}}(\cdot)\) only if \(f_{\theta_{k-1}}(\cdot)\) converges, BAN suffers from the sequential training and it severely reduces the training efficiency. 
 
 % aaai
 To make the teacher reap the benefits of the soft knowledge, \cite{yang2019aaai-second} emphasizes the importance of high-quality secondary information in prediction distribution,
 and Top Score Difference (TSD) regularization is proposed to make the prediction distribution less peaked to the primary class of the input samples.
 % ours
 Differently, to avoid sequential training, we propose an alternative method that makes use of the student prediction distribution of each task. Concretely, we add a feedback path from the student to the teacher and mutually regularize (MR) both the teacher and the student with the soft prediction from each other. 
 {
 % \color{blue}
 Besides the student is learned by Eqn. \ref{eq6} ($k$=1), the well-trained teacher network is further fine-tuned by}
 \begin{equation}
    \label{eq8}
    \mathcal{L}_{MR}  =
    \lambda_2\tau^2(\DJS(
    \hat{\mathcal{Y}}_{q}^{\theta_1}, \hat{\mathcal{Y}}_{q}^{\theta_0}
    )).
 \end{equation}
 
 % benefits of MR
 As improvements on unseen domains and TSD analysis shown in the ablation study, $\mathcal{L}_{MR}$ reliably counteracts the overfitting of the teacher. Since \(N_{w}\) classes are randomly selected for each FSC task, the true categories corresponding to the pseudo-label vary on different tasks. This allows the teacher to learn meaningful cross-category information from samples instead of remembering the pseudo-label. Besides, the student can leverage the one-hot label (in Eqn. \ref{eq6}) to ensure that both teachers and students are on the correct updating directions, resulting in non-degenerate solutions.
 
 \textbf{Design Choices.} 
 There are several potential variants to regularize the teacher network, including using both classification loss and $\mathcal{L}_{MR}$. 
 However, our goal is to make the teacher a regulator to provide soft knowledge and guide the approximate training direction such that the overfitting will not be passed to the students. 
 Updating the teacher with cross-entropy loss may retain overfitting. Moreover, applying $\mathcal{L}_{MR}$ to intermediate network layers is also possible, but it is computationally complex and hard to find the perfect design. Therefore, we choose a simple way to perform $\mathcal{L}_{MR}$ in the output space.

 \subsection{The Mismatched Teachers}
 \label{4.2}
 One core issue in DG-FSC is that we cannot get access to statistics from the target domain during training. Therefore, the reasonable way to improve the performance of an FSC model on \textit{unseen domains} is to improve the robustness and make it produce more stable predictions on \textit{various seen domains}.

 To formulate a training scheme that is similar to the test phase on unseen domains, inspired by the recent work \cite{li2019episodic_random}, we train the student in a way that exposes it to domain shift, making it robust to the mismatched source domain on which the current teacher is trained. 
 {
 % \color{blue}
 Concretely, for each source domain \(\mathcal{D}_i\), we train a teacher network using the training data of \(\mathcal{D}_i\) via Eqn. \ref{eq5},}
 and we denote the teacher obtained on \(\mathcal{D}_i\) as \(f_{\theta_0^*}^{D_i}(\cdot)\). In each iteration, for a task \(\mathcal{T}\) sampled from \(\mathcal{D}_i\), the student is updated in the same way as Eqn. \ref{eq6} but the teacher is obtained from a different domain \(\mathcal{D}_j\) that has never seen \(\mathcal{D}_i\) before. 
 We update the student using the ground truth and the mismatched soft outputs of \(f_{\theta_0^*}^{D_j}(\cdot)\), where \(i\neq j\), as $\mathcal{L}_{MM}$: 
  \begin{equation}
    \label{L_mm}
    \mathcal{L}_{MM} = \lambda_1\mathcal{L}_{ce}(\mathcal{Y}_q, \hat{\mathcal{Y}}_{q}^{\theta_1}) + \lambda_3\tau^2\DJS(
    \hat{\mathcal{Y}}_{q}^{\theta_1}, 
    \hat{\mathcal{Y}}_{q}^{(\theta_0, D_{j})}).
  \end{equation}

 How can a mismatched teacher (MM) help DG-FSC? Our insight is, if the student can be adapted to predict accurately on tasks from domain \(\mathcal{D}_i\) while guided by a mismatched teacher obtained on domain \(\mathcal{D}_j\) ($i \neq j$), then its robustness to domain-shift in the testing phase has increased. As minima quality analysis \cite{li2019episodic_random} shown in Sec. \ref{sec5}, $\mathcal{L}_{MM}$ improves domain-robustness compared to the baseline model. We further note that $\mathcal{L}_{MM}$ improves the generalization by a large margin on fine-grained unseen domains.
 
 \textbf{Design Choices.} In each task, the teacher is randomly selected which is mismatched to the domain of the current task. Compared to the teacher from the same source domain applied conventionally, the student \(f_{\theta_1}(\cdot)\) is penalized for the wrong prediction given the mismatched teacher that performs poorly on the current source domain. To minimize the total loss, the student model must learn to solve the task from the correct labels, but regularized by the teacher that is \textit{under domain-shift}. In the ablation study, we show the separate benefit of $\mathcal{L}_{MM}$ that it outperforms the baseline BAN consistently.
  %%%%%%%%%%%%%%%%%%%%%%%%%%%%
% \setlength{\tabcolsep}{1 mm}
% \renewcommand{\arraystretch}{1}
 % Table 3
 \begin{table*}[ht]
    \scriptsize
    \centering
    \caption{Meta-test accuracy (\%) for DG-FSC. Models are trained on miniImageNet and tested on various unseen domains.
    Note that in the following experiments there is only one domain involved in training, therefore only $\mathcal{L}_{MR}$ in FS-BAN is used. However, we show that FS-BAN still can improve the baseline FSC models consistently.
    }
    {
    \begin{adjustbox}{width=0.99\textwidth,center}
    \begin{tabular}{l c c c c c c c c}
         \toprule
            \multirow{2}{4em}{{Method}}
            & \multicolumn{2}{c}{{miniImageNet} $\mapsto$  {CUB}} 
            & \multicolumn{2}{c}{{miniImageNet} $\mapsto$ {Cars}} 
            & \multicolumn{2}{c}{{miniImageNet} $\mapsto$ {Places}} 
            & \multicolumn{2}{c}{{miniImageNet} $\mapsto$ {Plantae}} \\ \cline{2-9}
            & {{5-Way 1-Shot}} & {{5-Way 5-Shot}} & {{5-Way 1-Shot}} & {{5-Way 5-Shot}} & {{5-Way 1-Shot}} & {{5-Way 5-Shot}} & {{5-Way 1-Shot}} & {{5-Way 5-Shot}}\\
            \hline

            MatchingNet \cite{vinyals2016matching}
            & \(35.89 \pm 0.51\) & \(51.37 \pm 0.77\)
            & \bm{$30.77 \pm 0.68$} & \(38.99 \pm 0.64\)
            & \(49.86 \pm 0.79\) & \(63.16 \pm 0.77\)
            & \(32.70 \pm 0.60\) & \(46.53 \pm 0.68\)  \\

            +FT \cite{tseng2020cross}
            & \(36.64 \pm 0.53\) & \(55.23 \pm 0.83\)
            & \(29.82 \pm 0.44\) & \bm{$41.24 \pm 0.65$}
            & \(51.07 \pm 0.72\) & \(64.55 \pm 0.75\)
            & \(34.48 \pm 0.50\) & \(41.69 \pm 0.63\) \\
            
            +Baseline BAN (Our) 
            & {$36.47 \pm 0.53$} & {$51.07 \pm 0.58$}
            & {$28.71 \pm 0.43$} & {$37.29 \pm 0.49$}
            & {$51.05 \pm 0.70$} & {$64.19 \pm 0.61$}
            & {$35.01 \pm 0.53$} & {$46.78 \pm 0.53$} \\

            +FS-BAN (Our)
            & \bm{$41.03 \pm 0.58$} & \bm{$55.54 \pm 0.56$}
            & {$30.38 \pm 0.49$} & {$40.75 \pm 0.54$}
            & \bm{$53.88 \pm 0.66$} & \bm{$68.55 \pm 0.55$}
            & \bm{$36.05 \pm 0.53$} & \bm{$50.68 \pm 0.51$} \\
            \hline
            
            RelationNet \cite{sung2018relationnet}
            & \(42.44 \pm 0.77\) & \(57.77 \pm 0.69\)
            & \(29.11 \pm 0.60\) & \(37.33 \pm 0.68\)
            & \(48.64 \pm 0.85\) & \(63.32 \pm 0.76\)
            & \(33.17 \pm 0.64\) & \(44.00 \pm 0.60\)
            \\
            
            +FT \cite{tseng2020cross} 
            & \(44.07 \pm 0.77\) & \(59.46 \pm 0.71\)
            & \(28.63 \pm 0.59\) & \(39.91 \pm 0.69\)
            & \(50.68 \pm 0.87\) & \(66.28 \pm 0.72\)
            & \(33.14 \pm 0.62\) & \(45.08 \pm 0.59\)
            \\
            
            +LRP \cite{sun2020explanation}
            & \(42.44 \pm 0.41\) & \(59.30 \pm 0.40\)
            & \(29.65 \pm 0.33\) & \(39.19 \pm 0.38\)
            & \(50.59 \pm 0.46\) & \(66.90 \pm 0.40\)
            & \(34.80 \pm 0.37\) & \bm{$48.09 \pm 0.35$}
            \\
            
            +{Baseline BAN} (Our)
            & {$43.35 \pm 0.60$} & {$60.79 \pm 0.55$}
            & {$29.71 \pm 0.46$} & {$39.27 \pm 0.53$}
            & {$51.30 \pm 0.68$} & {$67.62 \pm 0.54$}
            & {$33.81 \pm 0.51$} & {$46.26 \pm 0.51$} \\
            
            +{FS-BAN} (Our)
            & \bm{$44.41 \pm 0.60$} & \bm{$61.31 \pm 0.55$}
            & \bm{$30.80 \pm 0.49$} & \bm{$40.47 \pm 0.54$}
            & \bm{$53.97 \pm 0.72$} & \bm{$70.21 \pm 0.56$}
            & \bm{$35.36 \pm 0.54$} & {$47.95 \pm 0.54$} \\
            \hline

            GNN \cite{garcia2017gnn}
            & \(45.69 \pm 0.68\)  & \(62.25 \pm 0.65\)
            & \(31.79 \pm 0.51\) & \(44.28 \pm 0.63\)
            & \(53.10 \pm 0.80\) & \(70.84 \pm 0.65\)
            & \(35.60 \pm 0.56\) & \(52.53 \pm 0.59\) \\

            +{FT} \cite{tseng2020cross}
            & \(47.47 \pm 0.75\) & \(66.98 \pm 0.68\)
            & \(31.61 \pm 0.53\) & \(44.90 \pm 0.64\)
            & \(53.77 \pm 0.79\) & \(73.94 \pm 0.67\)
            & \(35.95 \pm 0.58\) & \(53.85 \pm 0.62\) \\
         
            +LRP \cite{sun2020explanation}
            & \(48.29 \pm 0.51\) & \(64.44 \pm 0.48\)
            & \(32.78 \pm 0.39\) & \(46.20 \pm 0.46\)
            & \(54.83 \pm 0.56\) & \(74.45 \pm 0.47\)
            & \(37.49 \pm 0.43\) & \(54.46 \pm 0.46\) \\
            
            +{Baseline BAN} (Our)
            & {$47.08 \pm 0.70$} & {$68.30 \pm 0.68$}
            & {$32.22 \pm 0.56$} & {$44.65 \pm 0.63$}
            & {$54.82 \pm 0.80$} & {$74.94 \pm 0.66$}
            & {$36.71 \pm 0.63$} & {$55.34 \pm 0.63$} \\

            +{FS-BAN} (Our)
            & \bm{$50.04 \pm 0.76$} & \bm{$69.83 \pm 0.66$}
            & \bm{$33.21 \pm 0.60$} & \bm{$46.48 \pm 0.66$}
            & \bm{$59.70 \pm 0.84$} & \bm{$75.91 \pm 0.65$}
            & \bm{$38.87 \pm 0.64$} & \bm{$56.09 \pm 0.66$} \\
        \bottomrule
    \end{tabular}
    \end{adjustbox}
    \vspace{-3 mm}
    }
    \label{table3}
\end{table*}

 % Table 4
 \begin{table*}[t]
     \centering
     \caption{Conventional FSC results with model trained and tested solely on miniImageNet or tieredImageNet. Follow \cite{tseng2020cross}, we use ResNet-10 as the backbone and we show that our method can surpass the state-of-the-art methods with fewer parameters.}
    \begin{adjustbox}{width=0.7\textwidth,center}
    \begin{tabular}{l c c c c c}
    \toprule
        \multirow{2}{4em}{Method}
        & \multirow{2}{4em}{Backbone}
        % & \multirow{2}{8em}{\# of Parameters} 
        & \multicolumn{2}{c}{{miniImageNet} (base $\mapsto$ novel)}
        & \multicolumn{2}{c}{{tieredImageNet} (base $\mapsto$ novel) } \\\cline{3-6}
        
        & &
        {5-Way 1-Shot} & {5-Way 5-Shot} & {5-Way 1-Shot} & {5-Way 5-Shot} \\ \hline
        % MAML \cite{finn2017maml} & & Conv-4 & {$48.70 \pm 1.84$} &  {$63.11 \pm 0.92$}\\
        % ProtoNet \cite{snell2017prototypical} & & Conv-4 & {$49.42 \pm 0.78 $} & {$68.20 \pm 0.66$}\\
        % RelationNet \cite{sung2018relationnet} & & Conv-4 & {$50.44 \pm 0.82$} & {$65.32 \pm 0.70$}\\\hline
        % Transduction \cite{liu2018tpn} & & Conv-4 & \bm{$55.51 \pm 0.86$} & \bm{$69.86 \pm 0.65$} & \bm{$59.91 \pm 0.94$} & \bm{$73.30 \pm 0.75$}\\
        TADAM \cite{oreshkin2018tadam} & ResNet-12 & {$58.50 \pm 0.30$} & {$76.70 \pm 0.30$} & {$-$} & {$-$}\\ 
        MTL \cite{sun2019meta-transfer} & ResNet-12 & {$61.20 \pm 1.80$} & {$75.50 \pm 0.80$}\\
        CAN \cite{hou2019CAN} & ResNet-12 & {$63.85 \pm 0.48$} &  {$79.44 \pm 0.34$}  & {$69.89 \pm 0.34$} & {$84.23 \pm 0.37$} \\
        MetaOptNet \cite{lee2019meta-opt} & ResNet-12 & {$64.09 \pm 0.62$} & {$80.00 \pm 0.45$} & {$65.99 \pm 0.72$} & {$81.56 \pm 0.53$}\\
        Neg-Cosine \cite{liu2020negative-margin} & ResNet-12 & \(63.85 \pm 0.81\) & {$81.57 \pm 0.43$} & {$-$} & {$-$}\\
        RFS \cite{tian2020rethinkingsd-fsc} & ResNet-12 & {$64.82 \pm 0.60$} & {$82.14 \pm 0.56$} & \bm{$71.52 \pm 0.69$} & {$86.03 \pm 0.49$}\\
        % DeepEMD \cite{zhang2020deepemd} & & ResNet-12 & \bm{$65.91 \pm 0.82$} & \bm{$82.41 \pm 0.56$} 
        % & {$71.16 \pm 0.87$} & {$86.03 \pm 0.58$}\\
        
        \hline
        % Prediction \cite{qiao2018prediction} & & WRN-28-10 & {$59.60 \pm 0.41$} & {$73.74 \pm 0.19$}\\
        % Dynamic \cite{gidaris2018dynamic} & WRN-28-10 & {$60.06 \pm 0.14$} & {$76.39 \pm 0.11$} & {$67.92 \pm 0.16$} & \bm{$83.10 \pm 0.12$}\\
        LEO \cite{rusu2018-leo} & WRN-28-10 & {$61.76 \pm 0.08$} & {$77.59 \pm 0.12$} & {$66.33 \pm 0.05$} & {$81.44 \pm 0.09$}\\
        DAE-GNN \cite{gidaris2019dae-gnn} & WRN-28-10 & {$62.96 \pm 0.15$} & {$78.85 \pm 0.10$} & {$68.18 \pm 0.16$} & {$83.09 \pm 0.12$}\\
        AWGIM \cite{guo2020awgim} & WRN-28-10 & {$63.12 \pm 0.08$} & {$78.40 \pm 0.11$} & {$67.69 \pm 0.11$} & {$82.82 \pm 0.13$}\\\hline
        % MatchingNet & ResNet-10 & {$59.10 \pm 0.64$} & {$70.96 \pm 0.65$} & {$60.59 \pm 0.55$} & {$76.22 \pm 0.77$}\\ 
        % +MR (Our) & ResNet-10 & \bm{$62.28 \pm 0.64$} & \bm{$75.31 \pm 0.51$} & \bm{$62.88 \pm 0.75$} & \bm{$ 77.49 \pm 0.49 $}\\ \hline
        % RelationNet & ResNet-10 & {$57.80 \pm 0.88 $} & {$71.00 \pm 0.69$}  & {$61.49 \pm 0.59$} & {$78.98 \pm 0.53$}\\ 
        % +MR (Our) & ResNet-10 & \bm{$62.04 \pm 0.69$} & \bm{$76.69 \pm 0.52$} & \bm{$63.52 \pm 0.89$} & \bm{$80.12 \pm 0.32$}\\\hline
        GNN \cite{garcia2017gnn} & ResNet-10 & {$60.77 \pm 0.75$} & {$80.87 \pm 0.56$} & {$66.37 \pm 1.09$} & {$85.79 \pm 0.51$}\\
        +FT \cite{tseng2020cross} & ResNet-10 & {$66.32 \pm 0.80$} & {$81.98 \pm 0.55$} & {$-$} & {$-$}\\
        +FS-BAN (Our) & ResNet-10 &  \bm{$68.82 \pm 0.78$} & \bm{$84.89 \pm 0.50$} & {$70.55 \pm 0.78$} & \bm{$88.80 \pm 0.26$}\\
    \bottomrule
    \end{tabular}
    \end{adjustbox}
    \vspace{-4 mm}
    \label{table4}
 \end{table*}
 %%%%%%%%%%%%%%%%%%%%%%%%%%%%
 % MCT
\subsection{Meta-Control the Temperature}
\label{4.3}
 A fixed temperature (\(\eg, \tau=4\)) is often applied in the BAN and KD training process to soften the prediction probability distribution. Therefore, the student model can learn the inter-class relationships predicted by the well-trained teacher network. However, in the DG-FSC setup, the fixed temperature is applied to various source domains, of which there may be large differences and it leads to sub-optimal performance. For example, in some tasks, a higher temperature may result in less difference between classes. We propose to use meta-learned temperature tuning on different source domains. Our idea is that with the adaptively tuned $\tau$ that is proper to various seen domains the student can learn appropriate inter-class knowledge and improve the performance on unseen domains. 
 
 Instead of directly updating $\tau$, we propose a meta-learning scheme \cite{finn2017maml, gong2022meta, gong2022system, abdollahzadeh2021revisit} to efficiently tune the temperature (MCT):
 % MCT training process
 In iteration \(t\), we sample two subtasks from two different source domains: \(\mathcal{T}_{1} \in {D_{i}}\), and \(\mathcal{T}_{2} \in {D_{j}}\). Firstly, we update the student network \(f_{(\theta_1, t)}(\cdot)\) on \(\mathcal{T}_{1}\), given a pre-determined \(\tau_t\).
 Then, for task \(\mathcal{T}_2\), we fix the weights of the student \(f_{(\theta_1, t+1)}(\cdot)\) and evaluate the effectiveness of the temperature \(\tau_{t}\) that is applied to train the student on \(\mathcal{T}_1\), by testing the performance of \(f_{(\theta_1, t+1)}(\cdot)\). In this step, we use only cross-entropy loss (\(\tau\)=1), which is the same as the testing phase. The temperature \(\tau_{t+1}\) is obtained by evaluating \(f_{(\theta_1, t+1)}(\cdot)\) on query set \(\mathcal{Q}_2=\{\mathcal{X}_{(q,2)}, \mathcal{Y}_{(q,2)}\}\) of \(\mathcal{T}_2\):
 \begin{equation}
    \mathcal{L}_{MCT}=\mathcal{L}_{ce}(\mathcal{Y}_{(q,2)}, \hat{\mathcal{Y}}_{(q,2)}^{(\theta_{1}, t+1)}).
 \end{equation}
 \(\tau_{t+1}\) is used in the next iteration \textit{t+1}. With the adaptively fine-tuned temperature, we obtain a meta-learned hyperparameter that is trained to adapt to diverse domains. 
   
 \textbf{Design Choices.} Potentially, we have several ways to tune the temperature: 
 (i) The simplest way is that the temperature is updated directly as a normal learnable parameter in episodic training.
 (ii) We update the student on \(\mathcal{T}_1\) and evaluate the effectiveness of the temperature on \(\mathcal{T}_2\), but both tasks are from the same domain, \ie, \(\mathcal{T}_1, \mathcal{T}_2 \in \mathcal{D}_1\). 
 (iii) (Proposed $\mathcal{L}_{MCT}$ in FS-BAN) We update the student on \(\mathcal{T}_1\) and evaluate the effectiveness of the temperature on \(\mathcal{T}_2\), and the two tasks are from different source domains \ie, \(\mathcal{T}_1 \in \mathcal{D}_{i}, \mathcal{T}_2 \in \mathcal{D}_{j}\), as Figure \ref{FS-BAN}.
 In Sec. \ref{sec5}, 
 the temperature with setup (iii) converges gradually in the training process and gains better performance in the evaluation stage, indicating that we find a temperature suitable to diverse domains and training tasks. Therefore, we choose this setup for FS-BAN.

\subsection{Multi-Task Learning Objectives}
 The final learning objective of FS-BAN is:
  \begin{equation}
    \mathcal{L} = \mathcal{L}_{MR} + \mathcal{L}_{MM} + \mathcal{L}_{MCT}.
 \end{equation}

 In the next, we conduct comprehensive experiments to evaluate the effectiveness of FS-BAN on public datasets with popular FSC models as baselines. Detailed ablation studies and analyses are performed both qualitatively and quantitatively.

% section 6 experiments
\section{Experiments}
\label{sec5}
 
 In this section, we discuss the experiment settings and evaluate the proposed FS-BAN on six publicly available datasets with three popular metric baseline FSC models. We also conduct detailed ablation studies.
 
 \subsection{Datasets} 

 We evaluate the proposed FS-BAN on six publicly available datasets: miniImageNet \cite{ravi2016optimization}, tierdImageNet \cite{ren18tieredImagenet}, Caltech-UCSD Birds 200 (CUB) \cite{welinder2010caltech}, Stanford Cars (Cars) \cite{fei-fei-li2013-cars}, Places \cite{zhou2017places} and Plantae \cite{van2018plantae}. We follow the dataset split protocol as previous work \cite{tseng2020cross} for a fair comparison, and we summarize it in Table \ref{table_dataset}. In the meta-training phase, we use the standard data augmentation skills, including image jittering, random crop, random horizontal flip, and normalization for better generalization. In the meta-valid and meta-test stages, we do not use data augmentation.

 %%%%%%%%%% %%%%%%%%%%%%%%%%%
 % ablation study of FS-BAN
\begin{table*}[t]
    \centering
    \caption{
    Ablation study of FS-BAN with meta-test accuracy (\%).
    Model is trained on several seen source domains and evaluated with 5-Way 5-Shot tasks on the leave-one-out selected unseen domain.
    {
    % \color{blue}
    We show that each proposed component in FS-BAN improves the baseline models separately and they are complementary to each other.
    }
    }
    \label{table_ablation_study}
    \begin{adjustbox}{width=0.98\textwidth}
    \begin{tabular}{l cccccccc}
        \toprule
             5-Way 5-Shot & & $\mathcal{L}_{MR}$ & $\mathcal{L}_{MCT}$ & $\mathcal{L}_{MM}$ & {{All$\backslash$ \{CUB\}$\mapsto$CUB}}  & {{All$\backslash$ \{Cars\}$\mapsto$Cars}}  & {{All$\backslash$ \{Places\}$\mapsto$Places}}  & {{All$\backslash$ \{Plantae\}$\mapsto$Plantae}}  \\
            \hline
            
            MatchingNet \cite{vinyals2016matching} &  & - & - & -
            & {$51.92 \pm 0.80$}
            & {$39.87 \pm 0.51$}
            & {$61.82 \pm 0.57$}
            & {$47.29 \pm 0.51$} \\
    
            \quad  & FT\cite{tseng2020cross} & - & - & -
            & \(56.29 \pm 0.80\)
            & \(39.58 \pm 0.54\)
            & \(62.32 \pm 0.58\)
            & \(46.48 \pm 0.52\)\\
            
            \quad & LFT \cite{tseng2020cross} & - & - & - &
            {$61.41 \pm 0.57$}
            & {$43.08 \pm 0.55$}
            & \(64.99 \pm 0.59\)
            & \(48.32 \pm 0.57\)\\
            \hline
            
             & Baseline BAN & - & - & -
             & {$53.47 \pm 0.58$}
             & {$39.60 \pm 0.51$}
             & {$62.37 \pm 0.60 $}
             & {$48.42 \pm 0.57$} \\
             
             &  & \checkmark & - & -
             & {$59.75 \pm 0.56$}
             & {$42.03 \pm 0.55$}
             & {$69.34 \pm 0.57$}
             & {$54.61 \pm 0.58$} \\
             
             &  & - & \checkmark & -
             & {$55.97 \pm 0.59$}
             & {$41.97 \pm 0.55$}
             & {$64.37 \pm 0.58$}
             & {$50.61 \pm 0.59$} \\
             
             &  & - & - &  \checkmark
             & {$57.28 \pm 0.58$}
             & {$44.83 \pm 0.59$}
             & {$68.21 \pm 0.57$}
             & {$55.35 \pm 0.55$} \\
             
             &  & - & \checkmark  & \checkmark
             & {$58.25 \pm 0.59$}
             & {$42.91 \pm 0.56$}
             & {$66.22 \pm 0.55$}
             & {$51.99 \pm 0.54$} \\
             
             &  & \checkmark & \checkmark & -
             & {$61.64 \pm 0.59$}
             & {$42.18 \pm 0.56$}
             & {$69.80 \pm 0.58$}
             & \bm{$56.38 \pm 0.60$} \\
             
             & FS-BAN-lite & \checkmark & - & \checkmark
             & \bm{$62.83 \pm 0.59$}
             & {$42.44 \pm 0.56$}
             & {$69.84 \pm 0.55$}
             & {$54.87 \pm 0.57$} \\
             
             & FS-BAN & \checkmark & \checkmark & \checkmark
             & {$61.34 \pm 0.52$}
             & \bm{$45.01 \pm 0.57$}
             & \bm{$70.09 \pm 0.60$}
             & {$53.89 \pm 0.64$}\\
        \bottomrule
    \end{tabular}
    % \vspace{-5 mm}
    \end{adjustbox}
    \end{table*}

\begin{table}[t]
    \footnotesize
    \centering
    \caption{Collection of domains and the class split.}
    \label{table_dataset}
    \begin{adjustbox}{width=0.99 \columnwidth}
    \begin{tabular}{l|cccccc}
    \toprule
        \textbf{Domain} & {miniImageNet} & tieredImageNet & {CUB} & {Cars} & {Places}  & {Plantae}\\\hline
        {\# Training classes } & 64 & 351 & 100 & 98 & 183 & 100\\
        {\# Valid classes} & 16 & 97 & 50 & 49 & 91 & 50\\
        {\# Test classes} & 20 & 160 & 50 & 49 & 91 &  50\\
        % {Split Protocol} & Ravi \etal
        %  \cite{ravi2016optimization} & Ren \etal \cite{ren18tieredImagenet} & Hillard \etal  \cite{hilliard2018cub-split} & Tseng \etal \cite{tseng2020cross}& Tseng \etal \cite{tseng2020cross} & Tseng \etal \cite{tseng2020cross} \\
    \bottomrule
    \end{tabular}
    \end{adjustbox}
\end{table}

 %%%%%%%%%% %%%%%%%%%%%%%%%%%
 % ablation study of FS-BAN
\begin{table*}[t]
    \centering
    \caption{
    {
    % \color{blue}
    Meta-test accuracy (\%) with different implementation techniques for mutual regularization. 
    Model is trained on several seen source domains and evaluated on the leave-one-out selected unseen domain with 5-Way 5-Shot tasks.
    The feature encoders of all models are pre-trained on mini-ImageNet.
    }
    }
    \label{table_r1_lmr_lmm_ablation}
    % \begin{adjustbox}{width=0.99\textwidth}
    {
    % \color{blue}
    \begin{tabular}{p{6cm}<{\raggedright} | cccccccc}
        \toprule
            \textbf{Method} & {{All$\backslash$ \{CUB\}$\mapsto$CUB}} & {{All$\backslash$ \{Cars\}$\mapsto$Cars}}  & {{All$\backslash$ \{Places\}$\mapsto$Places}}  & {{All$\backslash$ \{Plantae\}$\mapsto$Plantae}}  \\
            \hline
            
            MatchingNet \cite{vinyals2016matching}
            & {$51.92 \pm 0.80$}
            & {$39.87 \pm 0.51$}
            & {$61.82 \pm 0.57$}
            & {$47.29 \pm 0.51$} \\
            
            + Baseline BAN
            & {$53.47 \pm 0.58$}
            & {$39.60 \pm 0.51$}
            & {$62.37 \pm 0.60 $}
            & {$48.42 \pm 0.57$} \\\hline

            + $\mathcal{L}_{MR}$ (w/o student warmup) & $55.98 \pm 0.54$ & $40.51 \pm 0.57$ & $64.84 \pm 0.60$ & $50.37 \pm 0.57$\\
            
            + $\mathcal{L}_{MR}$ (w/ student warmup) & $58.46 \pm 0.55$ & $41.82 \pm 0.55$ & $68.56 \pm 0.61$ & $52.77 \pm 0.54$
            \\
            
            + $\mathcal{L}_{MR}$ (w/ student warmup + reduced teacher $lr$)
            & \bm{$59.75 \pm 0.56$}
            & \bm{$42.03 \pm 0.55$}
            & \bm{$69.34 \pm 0.57$}
            & \bm{$54.61 \pm 0.58$} \\
            
            % + FS-BAN & 
            % & \bm{$61.34 \pm 0.52$}
            % & \bm{$45.01 \pm 0.57$}
            % & \bm{$70.09 \pm 0.60$}
            % & \bm{$53.89 \pm 0.64$}\\
        \bottomrule
    % \end{tabular}}
    % % \end{adjustbox}
    % {\color{blue}
    % \begin{adjustbox}{width=0.94\textwidth}
    % \begin{tabular}{p{6cm}<{\raggedright} | cccccccc}
        % \toprule
        %      \textbf{Method} & {{All$\backslash$ \{CUB\}$\mapsto$CUB}}  & {{All$\backslash$ \{Cars\}$\mapsto$Cars}}  & {{All$\backslash$ \{Places\}$\mapsto$Places}}  & {{All$\backslash$ \{Plantae\}$\mapsto$Plantae}}  \\
        %     \hline
            
        %     MatchingNet
        %     & {$51.92 \pm 0.80$}
        %     & {$39.87 \pm 0.51$}
        %     & {$61.82 \pm 0.57$}
        %     & {$47.29 \pm 0.51$} \\
            
        %     + Baseline BAN ($\mathcal{L}_{ce}$ + w/ matched teacher)
        %     & {$53.47 \pm 0.58$}
        %     & {$39.60 \pm 0.51$}
        %     & {$62.37 \pm 0.60 $}
        %     & {$48.42 \pm 0.57$} \\ 
        %     \hline
            
        %     + $\mathcal{L}_{MM}$ ($\mathcal{L}_{ce}$ + w/ noisy teacher)
        %     & {$ 52.83 \pm 0.54$}
        %     & {$ 40.22 \pm 0.53$}
        %     & {$ 62.83 \pm 0.58$}
        %     & {$ 50.64 \pm 0.58$} \\
            
        %     + $\mathcal{L}_{MM}$ ($\mathcal{L}_{ce}$ + w/ mismatched teacher)
        %     & \bm{$57.28 \pm 0.58$}
        %     & \bm{$44.83 \pm 0.59$}
        %     & \bm{$68.21 \pm 0.57$}
        %     & \bm{$55.35 \pm 0.55$} \\
        % \bottomrule
    \end{tabular}
    % \end{adjustbox}
    }
    \end{table*}
 % hyper-parameter coef
\begin{table*}[h]
    \footnotesize
    \centering
    \caption{Ablation study of coefficients for $\mathcal{L}_{MR}$ and $\mathcal{L}_{MM}$ of FS-BAN. We use the same experiment setup as Table \ref{table2}.}
    \begin{adjustbox}{width=0.85 \textwidth}
    \begin{tabular}{l cccccc}
        \toprule
            {{5-Way 5-Shot}} & \(\lambda_1\) & \(\lambda_2\) & 
            {{All$\backslash$ \{CUB\}$\mapsto$CUB}}  & {{All$\backslash$ \{Cars\}$\mapsto$Cars}}  & {{All$\backslash$ \{Places\}$\mapsto$Places}}  & {{All$\backslash$ \{Plantae\}$\mapsto$Plantae}}  \\
            \hline
            
            MatchingNet + $\mathcal{L}_{MR}$ & 1 & 0 & {$51.92 \pm 0.80$} & {$39.87 \pm 0.51$} & {$61.82 \pm 0.57$} & {$47.29 \pm 0.51$}\\
            
             & 1 & 0.2 & {$56.27 \pm 0.57$} & {$41.64 \pm 0.56$} & {$66.55 \pm 0.58$} & {$52.71 \pm 0.56$} \\
            
             & 1 & 0.5 & {$57.50 \pm 0.50$} & {$40.31 \pm 0.52$} & {$67.88 \pm 0.56$} & {$52.73 \pm 0.56$} \\
             
             & 1 & 0.8 & \bm{$59.75 \pm 0.56$} & \bm{$42.03 \pm 0.55$} & \bm{$69.34 \pm 0.57$} & \bm{$54.61 \pm 0.58$} \\
        \bottomrule
        \toprule
            {{5-Way 5-Shot}} & \(\lambda_1\) & \(\lambda_3\) & 
            {{All$\backslash$ \{CUB\}$\mapsto$CUB}}  & {{All$\backslash$ \{Cars\}$\mapsto$Cars}}  & {{All$\backslash$ \{Places\}$\mapsto$Places}}  & {{All$\backslash$ \{Plantae\}$\mapsto$Plantae}}  \\
                        \hline
            MatchingNet + $\mathcal{L}_{MM}$ & 1 & 0 & {$51.92 \pm 0.80$} & {$39.87 \pm 0.51$} & {$61.82 \pm 0.57$} & {$47.29 \pm 0.51$}\\

             & 1 & 0.2 & {$55.42 \pm 0.60$} & {$42.89 \pm 0.57$} & {$66.77 \pm 0.57$} & {$52.87 \pm 0.55$} \\
             & 1 & 0.5 & \bm{$57.28 \pm 0.58$} & {$44.83 \pm 0.59$} & \bm{$68.21 \pm 0.57$} & \bm{$55.35 \pm 0.55$} \\
             & 1 & 0.8 & {$55.32 \pm 0.57$} & \bm{$45.45 \pm 0.58$} & {$67.98 \pm 0.56$} & {$53.26 \pm 0.57$} \\
        \bottomrule
    \vspace{-3.5 mm}
    \end{tabular}
    \end{adjustbox}
    \label{table_coef}
\end{table*}

 %%%%%%%%%% %%%%%%%%%%%%%%%%%
 \subsection{Baseline Models}
 Since FS-BAN does not require additional learnable parameters and can be readily used to existing FSC methods, we apply FS-BAN to three popular metric-based FSC models to validate the effectiveness of FS-BAN: MatchingNet \cite{vinyals2016matching}, RelationNet \cite{sung2018relationnet} and Graph Neural Network (GNN) \cite{garcia2017gnn}. All these baseline models share the same feature extractor as the backbone network and only differ in the metric-based classifier head for prediction. 
 For other DG-FSC methods, we compare with \cite{tseng2020cross} that applies feature-wise transformation layers (LFT) to improve the generalization. We also compare with layer-wise relevance propagation (LRP, \cite{sun2020explanation}) and more state-of-the-art FSC models in both single domain and DG setups.
 
 \subsection{Experiment Setups} 
 For a fair comparison, we follow \cite{tseng2020cross} to assume there are multiple seen source domains in training. Nevertheless, to comprehensively evaluate different methods, in the main experiments, we perform three experiment setups: 
 \begin{enumerate}
     \item Models are trained on tasks of base classes of multiple seen source domains and tested on a target unseen domain. The source domains are selected from All=\{miniImageNet, CUB, Cars, Places, Plantae\}, \eg, All$\backslash$ \{CUB\}. The unseen domain for testing is the held-out domain during training, \eg, CUB.
     
     \item Models are trained on a single source domain, \ie, base classes of miniImageNet, and tested on novel classes of various unseen domains, \eg, All$\backslash$ \{miniImageNet\}.
     
     \item We further perform conventional FSC \cite{sung2018relationnet, guo2020awgim} experiments, where base classes for training and novel classes for evaluation are from the same domain (\eg, miniImageNet).  
 \end{enumerate}
 Note that in setups 2) and 3) there is only one source domain involved in training, therefore only $\mathcal{L}_{MR}$ in FS-BAN is applicable. However, we show that this partial FS-BAN can still outperform other methods.

 For all experiment setups, follow \cite{tseng2020cross, sun2020explanation}, we use ResNet-10 \cite{he2016resnet} as the backbone network for baseline models and our method. 
 We initialize the temperature with \(\tau=4\), and it is activated by a SoftPlus function to ensure it is non-negative:
 \begin{equation}
     \tau = \textnormal{SoftPlus}(\tau) = \ln{(1+e^{\tau})},
 \end{equation}
 {
 % \color{blue}
 where $\tau$ is updated in each iteration, as described in Sec. \ref{4.3}.
 }
 \subsection{
 % \color{blue}
 Implementation Details}
 We strictly follow the standard FSC setups \cite{rusu2018-leo, guo2020awgim, sung2018relationnet, snell2017prototypical}: either 5-Way 1-Shot or 5-Way 5-Shot tasks are sampled in training and testing stages. 
 In each task, we sample $N_{q}=16$ query images per category to compute the loss and accuracy. 
 We train FS-BAN with 800 epochs (100 tasks are sampled from a random source domain in each epoch). 
 We apply the Adam optimizer \cite{kingma2014adam} to train the models with default hyperparameters, \eg, learning rate 0.001.
 In the testing phase, we sample 1,000 tasks of novel classes from the unseen target domain in setup 1) and 2) and the same source domain in setup 3) for evaluation, respectively. 
 We select the model checkpoints with the best validation accuracy and 
 report the average accuracy
 on the test set with 95\% confidence interval.
 
 {
 % \color{blue}
 On the other hand, we follow the prior works \cite{tseng2020cross, rusu2018-leo} to pre-train the backbone model (ResNet-10 \cite{he2016resnet} feature encoder with a linear layer as the classifier) on 64 base classes of mini-ImageNet, by minimizing a standard cross-entropy loss, as Eqn. \ref{eq1}. 
 After that, we remove the classifier head and use the pre-trained backbone weights to initialize the student network for episodic training for DG-FSC. 
 Therefore, at the beginning of the meta-training stage, the student is equipped with a feature encoder that can extract discriminative features.
 We use this technique in all our experiments as it has been shown very useful in FSC in prior works \cite{tseng2020cross, rusu2018-leo, gidaris2018dynamic, lifchitz2019dense}.
 }
 
 \subsection{Experiment Results}
 The results of experiment setup 1), 2) and 3) are shown in Table \ref{table2}, Table \ref{table3}, and Table \ref{table4}, respectively. 
 In all setups, our proposed FS-BAN consistently improves the different baseline FSC models to state-of-the-art, presenting desirable performance on unseen domains. 
 {
 % \color{blue}
 Since there is no true label in episodic training for DG-FSC (\ie, samples are pseudo-labeled in different tasks), the results imply that our obtained models indeed learn generalizable knowledge that can help tackle different tasks on novel classes of unseen domains, as analysis in Table \ref{table-a3}, where we show the improved accuracy and lower top-score difference in the prediction distributions.
 
 Compared to the prior state-of-the-art method that introduces additional learnable parameters \cite{tseng2020cross}, our proposed FS-BAN can address unique issues of DG-FSC, including overfitting and domain shift, benefits the network generalizability on unseen target domains, and improves the performance without additional inference cost in the deployment stage. 
 We further note that in setup 3) where base classes for training and novel classes for evaluation are from the same domain (hence the domain gap is reduced), our method can still achieve considerable improvement consistently, even with fewer parameters of the backbone network, as Table \ref{table4}.
 }
 
 As we show in the ablation studies in the next, our proposed learning objectives in FS-BAN successfully address the unique challenges posed in DG-FSC, and the generalizability is greatly improved on unseen domains.
 
\subsection{Ablation Study}
\label{5.3}

 % We also perform ablation studies of the proposed FS-BAN. We show that the challenges of BAN in DG-FSC (as mentioned in Sec. \ref{3.3}) are effectively addressed.
%%%%%%%%%%%%%%%%%%
%%%%%%%%%%%%%%%%%%

 \textbf{Ablation study of learning objectives of FS-BAN.} 
 {
 % \color{blue}
 To evaluate the effectiveness of each individual component in the multi-task learning objectives of the proposed FS-BAN}, we conduct comprehensive ablation studies and observe the empirical performance of FS-BAN in the DG-FSC setup. 
 We use MatchingNet as the baseline model. We sample 5-Way 5-Shot tasks for training and evaluation, and other settings are the same as setup 1). 
 The results are shown in Table \ref{table_ablation_study}. 

 We show that each separate learning objective ($\mathcal{L}_{MR}$, $\mathcal{L_{MM}}$, $\mathcal{L}_{MCT}$) in FS-BAN improves the baseline models effectively and they are complimentary to each other. 
 On the other hand, the FS-BAN with the full  multi-task learning objectives achieves a good balance and performance on different unseen domains (the last row in Table \ref{table_ablation_study}).

 {
 % \color{blue}
 % mr
 In practice, to ensure that the feedback from the student prediction for ($\mathcal{L}_{MR}$) is reasonable and will not mislead the fine-tuning of the teacher network, \textit{esp.,} at the beginning of student training, we introduce a ``student warmup'' process to let the student train 10 epochs with randomly sampled tasks before its feedback to the well-trained teacher network. 
 We note that the backbone of the student model is pre-trained on mini-ImageNet training classes.
 We also reduce the learning rate (\textit{lr}) of the teacher network by a factor of 5 compared to that of the student network, such that the teacher model is only moderately updated. 
 In Table \ref{table_r1_lmr_lmm_ablation}, we study the impact of ``student warmup'' and ``reduced \textit{lr} for teacher'' for $\mathcal{L}_{MR}$, and we show that the adopted techniques can improve the performance for $\mathcal{L}_{MR}$ by a considerable margin.
 }

 % mm
 On the other hand, interestingly, when we only use FS-BAN with the mismatched teacher ($\mathcal{L}_{MM}$), the performance is better than all baselines and the state-of-the-art models with Cars being the unseen domain. This suggests improved generalization on the fine-grained datasets. 
 {\color{blue}
 % In Table \ref{table_r1_lmr_lmm_ablation} (Bottom), we additionally show that the proposed ($\mathcal{L}_{MM}$) can achieve better performance than training with a ``matched teacher'' or a simple regulator, where the teacher network is randomly initialized (dubbed ``noisy teacher''). We will discuss it further in Sec. \ref{sec_tsd}.
}

%%%%%%%%%%%%%%%%%%
%%%%%%%%%%%%%%%%%
 \textbf{Ablation study of coefficients in learning objectives.}
 % Here, we take a step further to dissect FS-BAN and study the advantages of each component. 
 We perform a grid search to select the coefficients of the loss objectives of $\lambda_2$ in $\mathcal{L}_{MR}$ and $\lambda_3$ in  $\mathcal{L}_{MM}$ in our work. We fix \(\lambda_1=1\) for cross-entropy loss  and tune \(\lambda_2\) for $\mathcal{L}_{MR}$ and \(\lambda_3\) for $\mathcal{L}_{MM}$. 
 In Table \ref{table_coef}, we show the accuracy with different choices of coefficients.
 The student is trained using setup 1) with 5-Way 5-Shot tasks. MatchingNet is the baseline model. 
 We observe that the performance of our proposed model is not very sensitive to different coefficients, and our methods can outperform the baseline method (where $\lambda_2=\lambda_3=0$) significantly. 
 {
 % \color{blue}
 Nevertheless, we note that it is possible that the mismatched teacher may harm the accuracy of the student, especially when such \textit{domain-shift} training in $\mathcal{L}_{MM}$ is over-emphasized. 
 Here, we found that  $\lambda_3=0.5$  is the best weight of $\mathcal{L}_{MM}$, which indicates that it is not over-emphasized.
 }
 Based on the empirical results in Table \ref{table_coef}, we choose \(\lambda_1=1, \lambda_2=0.8, \lambda_3=0.5\) as the coefficients in $\mathcal{L}_{MR}$ and $\mathcal{L}_{MM}$, which performs well in most experiments.

 \begin{figure}[t]
    \centering
    \includegraphics[width=0.99\columnwidth]{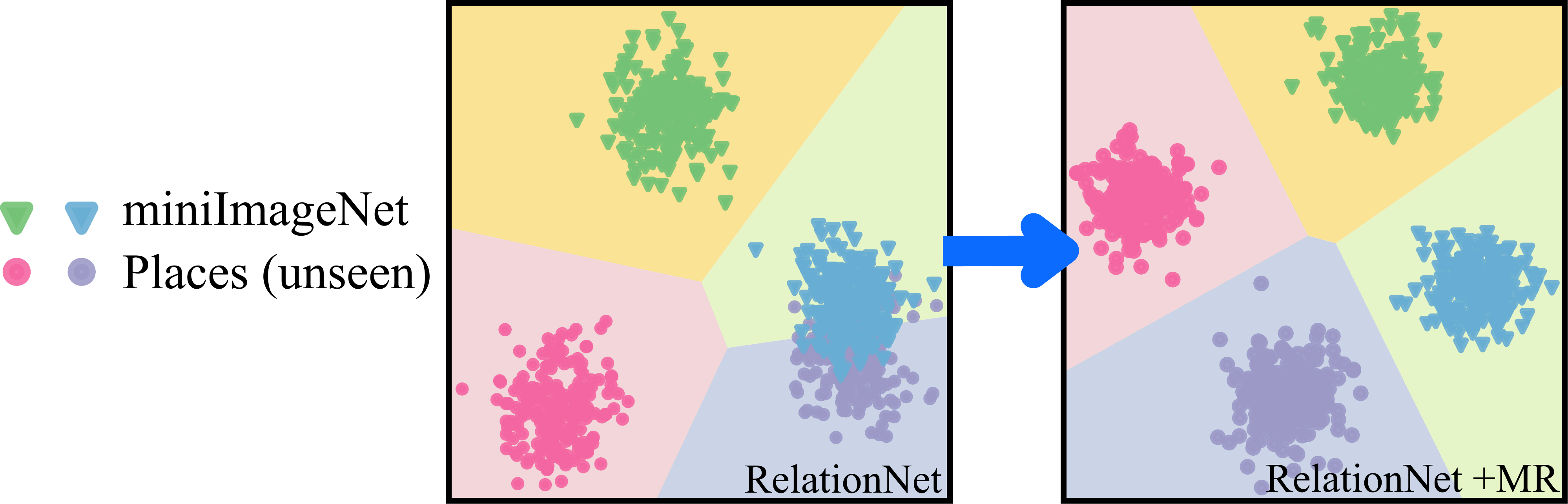}
    \caption{Qualitative evaluation of the class separation. We show the projection of novel class features of the first and second components of LDA. We sample 200 images from miniImageNet and Places (unseen) separately. It is clear that $\mathcal{L}_{MR}$ brings better decision boundaries for the DG-FSC setup. 
    }
    \label{lda_mr}
 \end{figure}
  \begin{table}[t]
    \centering
    \caption{Quantitative class separation evaluation for miniImageNet $\rightarrow$ Places. `RN' and `MN' indicate the RelationNet, and MatchingNet, respectively. 
    Following \cite{gold-icml2020-unraveling}, lower values correspond to better feature clustering of novel tasks.}
    \begin{adjustbox}{width=0.95\columnwidth}
    \begin{tabular}{c cc cc cc}
        \toprule
        Metric ($\downarrow$) & RN & RN+$\mathcal{L}_{MR}$ & MN & MN+$\mathcal{L}_{MR}$ & GNN & GNN+$\mathcal{L}_{MR}$ \\ \hline
        \(R_{FC}\) & 7.94 & \textbf{6.40} & 2.24 & \textbf{2.22} & 6.32 & \textbf{5.85} \\ \hline 
        \(R_{HV}\) & 1.81 & \textbf{1.73} & 1.78 & \textbf{1.59} & 1.75 & \textbf{1.71} \\
        \bottomrule
    \end{tabular}
    \end{adjustbox}
    \label{table5}
 \end{table}

 \begin{figure}[t]
    \centering
    {
        \includegraphics[width=0.7\columnwidth]{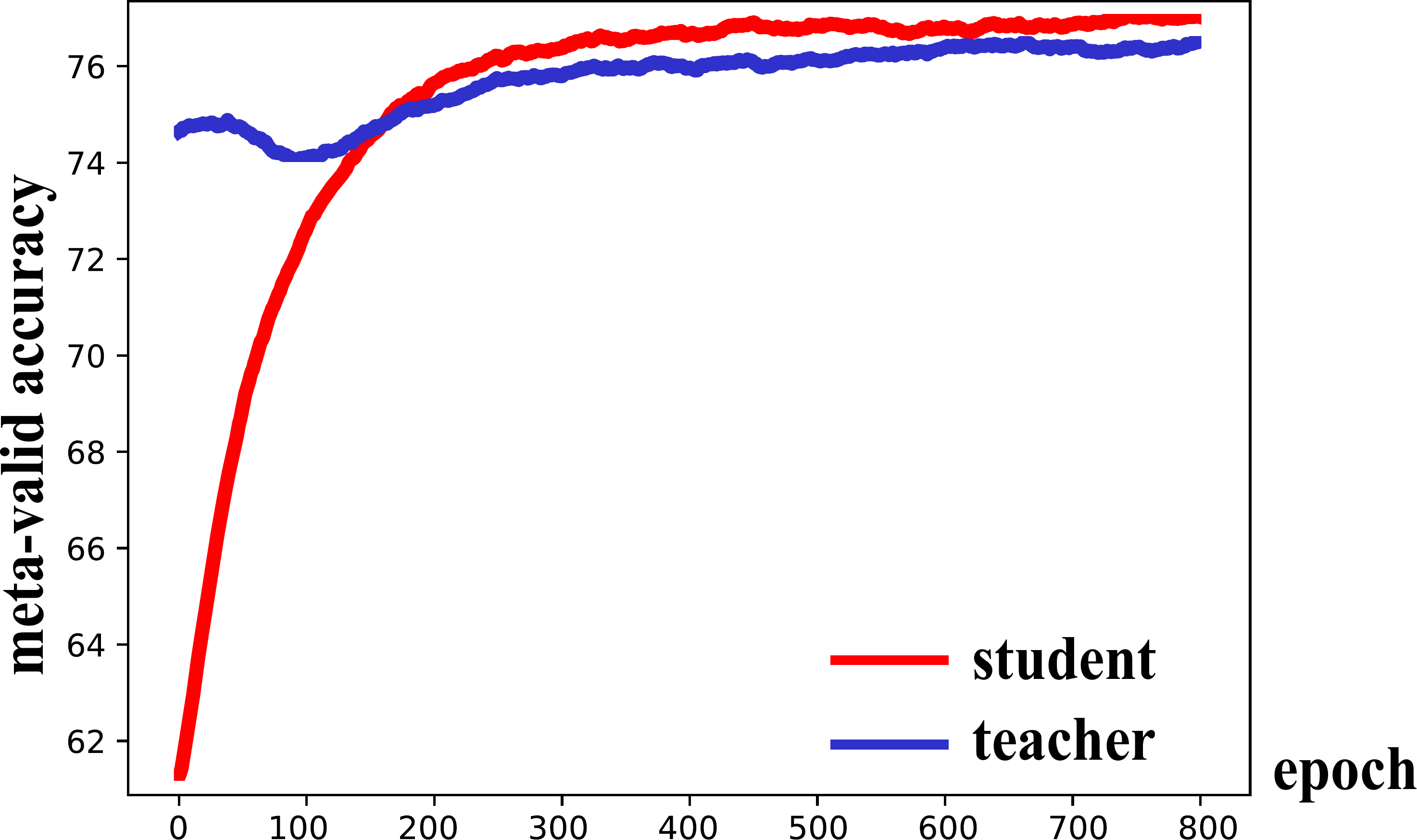}
    }
    \caption{
    {
    % \color{blue}
    Valid accuracy (\%) on novel classes of miniImageNet with only $\mathcal{L}_{MR}$ in FS-BAN.
    We show that, regularized by $\mathcal{L}_{MR}$, the well-trained teacher network can be continually improved and gain better performance on unseen novel classes. It in turn leads to a better student with improved generalizability that even outperforms the teacher network consistently.
    }
    }
    \label{mr_valid_acc}
 \end{figure}

\begin{figure}[t]
\centering
{
    \includegraphics[width=0.65\columnwidth]{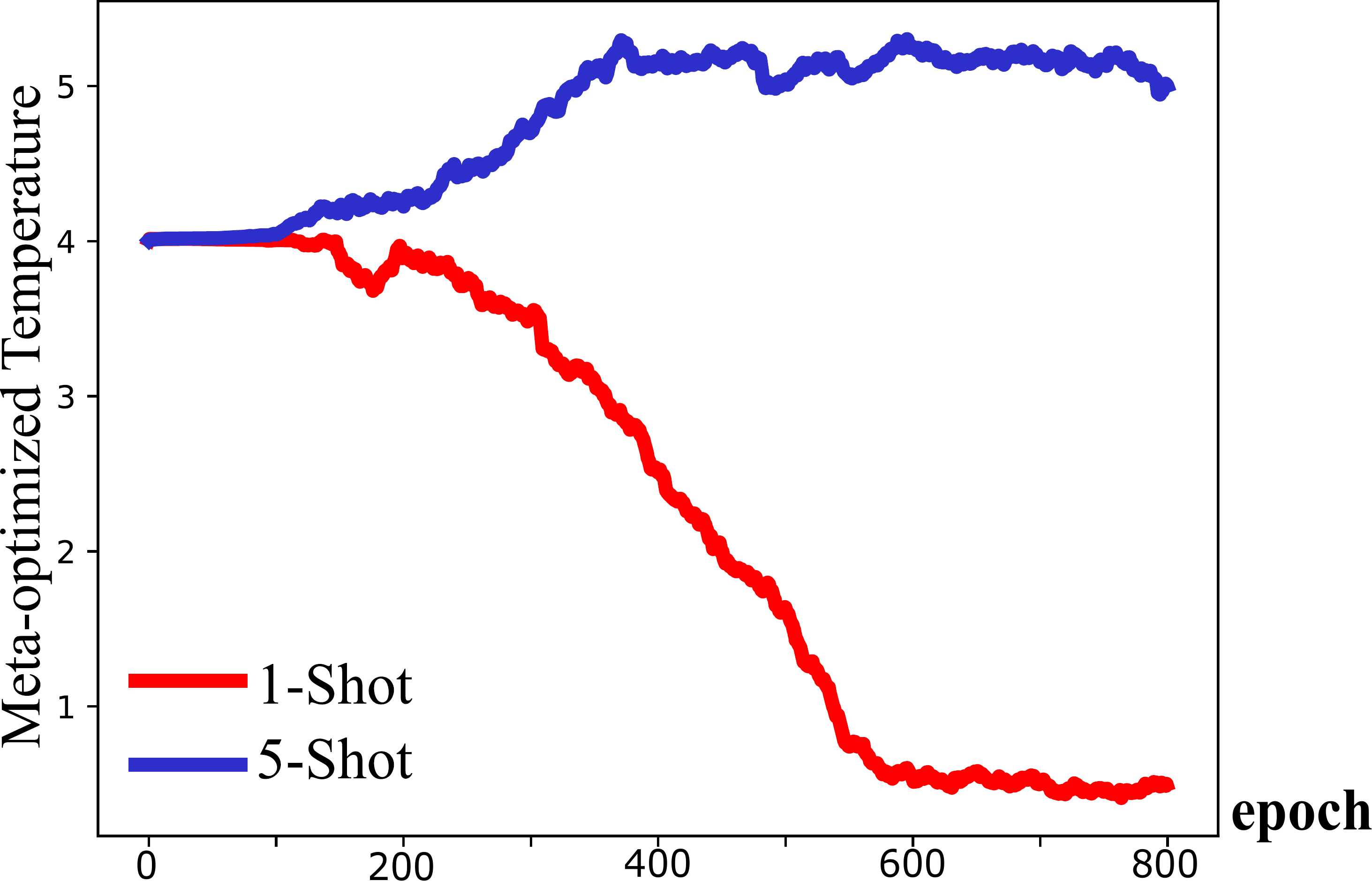}
}
\caption{
{
% \color{blue}
Visualization of the temperature ($\tau$) value during training with $\mathcal{L}_{MCT}$.
Compared to a fixed $\tau$, our proposed $\mathcal{L}_{MCT}$ can help find a proper temperature that is suitable for diverse domains. 
In Table \ref{mct_direct} we demonstrate the performance with different design choices of updating $\tau$ and show the effectiveness of the proposed $\mathcal{L}_{MCT}$.
}
}
\label{fig_temperature}
\vspace{-4 mm}
\end{figure}
\begin{table}[t]
    \centering
    \caption{Meta-test accuracy (\%) of DG-FSC with experiment setup 1) by updating the temperature with different design choices on 5-Way 5-Shot tasks.} 
    \begin{adjustbox}{width=0.98\columnwidth}
        \begin{tabular}{l lcc}
        \toprule
            Models & setting & 
            All$\backslash$\{CUB\}$\mapsto$CUB & 
            All$\backslash$\{Cars\}$\mapsto$Cars \\
            \cline{3-4}
            MatchingNet & no update $\tau$ & {$51.92 \pm 0.80$} & {$39.87 \pm 0.51$}\\
            \quad & directly update $\tau$ &  {$50.58 \pm 0.59$} & {$40.57 \pm 0.52$}\\
            \quad &  $ \mathcal{L}_{MCT}$ &  \bm{$55.97 \pm 0.59$} & \bm{$41.97 \pm 0.55$}\\\hline
            & &             All$\backslash$\{Places\}$\mapsto$Places & 
            All$\backslash$\{Plantae\}$\mapsto$Plantae\\ \cline{3-4}
            MatchingNet & no update $\tau$ & {$61.82 \pm 0.57$} & {$47.29 \pm 0.51$} \\
            & directly update $\tau$ & {$61.89 \pm 0.59$} & {$48.95 \pm 0.57$} \\
            \quad &  $ \mathcal{L}_{MCT}$ &  \bm{$64.37 \pm 0.58$} & \bm{$50.61 \pm 0.59$}\\
        \bottomrule
    \label{mct_direct}
    \end{tabular}
    \end{adjustbox}
\end{table}
 \begin{table}[t]
     \centering
     \small
     \caption{We train models on base classes of miniImageNet using 5-Way 5-Shot tasks. We present the average Top-score difference (TSD) of the prediction distribution of the teacher in training, and the performance of the student on novel classes from the unseen domain (Places) in testing.}
     \subtable[TSD of the teacher with M=3 in the training process.]
     {
        \begin{adjustbox}{width=0.70\columnwidth}
         \begin{tabular}{l c c}
         \toprule
             & Baseline BAN & Baseline BAN + $\mathcal{L}_{MR}$ \\\hline
            %   Training phase (teacher) & 2 &  0.74 &\textbf{0.60}     \\ \\\hline
              TSD & 0.78 & \textbf{0.64}     \\
         \bottomrule
         \end{tabular}
        \end{adjustbox}
     }
     \subtable[Meta-test accuracy (\%) on novel classes on different domains.]{
     \begin{adjustbox}{width=0.95 \columnwidth}
     \begin{tabular}{lcc}
     \toprule
           Dataset & Baseline BAN & Baseline BAN + $\mathcal{L}_{MR}$\\ \hline
           miniImageNet (base $\mapsto$ novel) & $73.79$ & \bm{$75.31$}\\
          miniImageNet$\mapsto$Places & $64.19$ & \bm{$68.55$}\\
     \bottomrule
     \end{tabular} 
     \end{adjustbox}
     }
     \label{table-a3}
     \vspace{-4mm}
 \end{table}
  \begin{table}[t]
     \centering
     \caption{The teacher is trained on miniImageNet and we evaluate its characteristics on different domains.}
     \begin{adjustbox}{width = 0.98\columnwidth}
     \begin{tabular}{lccccc}
     \toprule
         Teacher Model & miniImageNet & mini$\mapsto$CUB & mini$\mapsto$Cars & mini$\mapsto$Places & mini$\mapsto$Plantae \\\hline
          Accuracy (\%) & \textbf{70.96} & 51.37 & 38.99 & 63.16 & 46.53\\\hline
          TSD & \textbf{0.64} & 0.39 & 0.23 & 0.48 & 0.35\\
     \bottomrule
     \end{tabular}
     \end{adjustbox}
     \label{Table-MM}
 \end{table}
 
\begin{table*}[t]
    \centering
    \caption{
    {
    % \color{blue}
    Meta-test accuracy (\%) with different backbone networks of teacher models. Model is trained on several seen source domains and evaluated on the leave-one-out selected unseen domain with 5-Way 5-Shot tasks.
    }}
    \label{table_larger_backbone}
    \begin{adjustbox}{width=0.99\textwidth}
    {
    % \color{blue}
    \begin{tabular}{l ccccccccc}
        \toprule
             \textbf{Method} & \textbf{Backbone} & \textbf{Backbone of Teacher} &  
             {{All$\backslash$ \{CUB\}$\mapsto$CUB}}  & {{All$\backslash$ \{Cars\}$\mapsto$Cars}}  & {{All$\backslash$ \{Places\}$\mapsto$Places}}  & {{All$\backslash$ \{Plantae\}$\mapsto$Plantae}}  \\
            \hline
            
            MatchingNet & Conv-4 & - & $50.27 \pm 0.54$ & $37.75 \pm 0.52$ & $56.72 \pm 0.55$ & $43.22 \pm 0.59$ \\
            +FS-BAN     & Conv-4 & Conv-4 & $51.95 \pm 0.61$ & $43.20 \pm 0.56$ & $62.45 \pm 0.49$ & $44.28 \pm 0.61$  \\
            +FS-BAN     & Conv-4 & Conv-6 & \bm{$53.20 \pm 0.58$} & \bm{$44.95 \pm 0.52$} & \bm{$64.38 \pm 0.54$} & \bm{$48.14 \pm 0.60$} \\
            \midrule
            
            MatchingNet & ResNet-10 & -
            & {$51.92 \pm 0.80$}
            & {$39.87 \pm 0.51$}
            & {$61.82 \pm 0.57$}
            & {$47.29 \pm 0.51$} \\

            +FS-BAN & ResNet-10 & ResNet-10
            & {$61.34 \pm 0.52$}
            & {$45.01 \pm 0.57$}
            & {$70.09 \pm 0.60$}
            & {$53.89 \pm 0.64$}\\
            +FS-BAN & ResNet-10 & ResNet-18 & \bm{$61.58 \pm 0.55$} & \bm{$46.73 \pm 0.58$} & \bm{$70.31 \pm 0.61$} & \bm{$54.44 \pm 0.57$}
            \\
        \bottomrule
    \end{tabular}}
    \end{adjustbox}
    \end{table*}

 \begin{figure*}[t]
    \centering
    \includegraphics[width=0.245\textwidth]{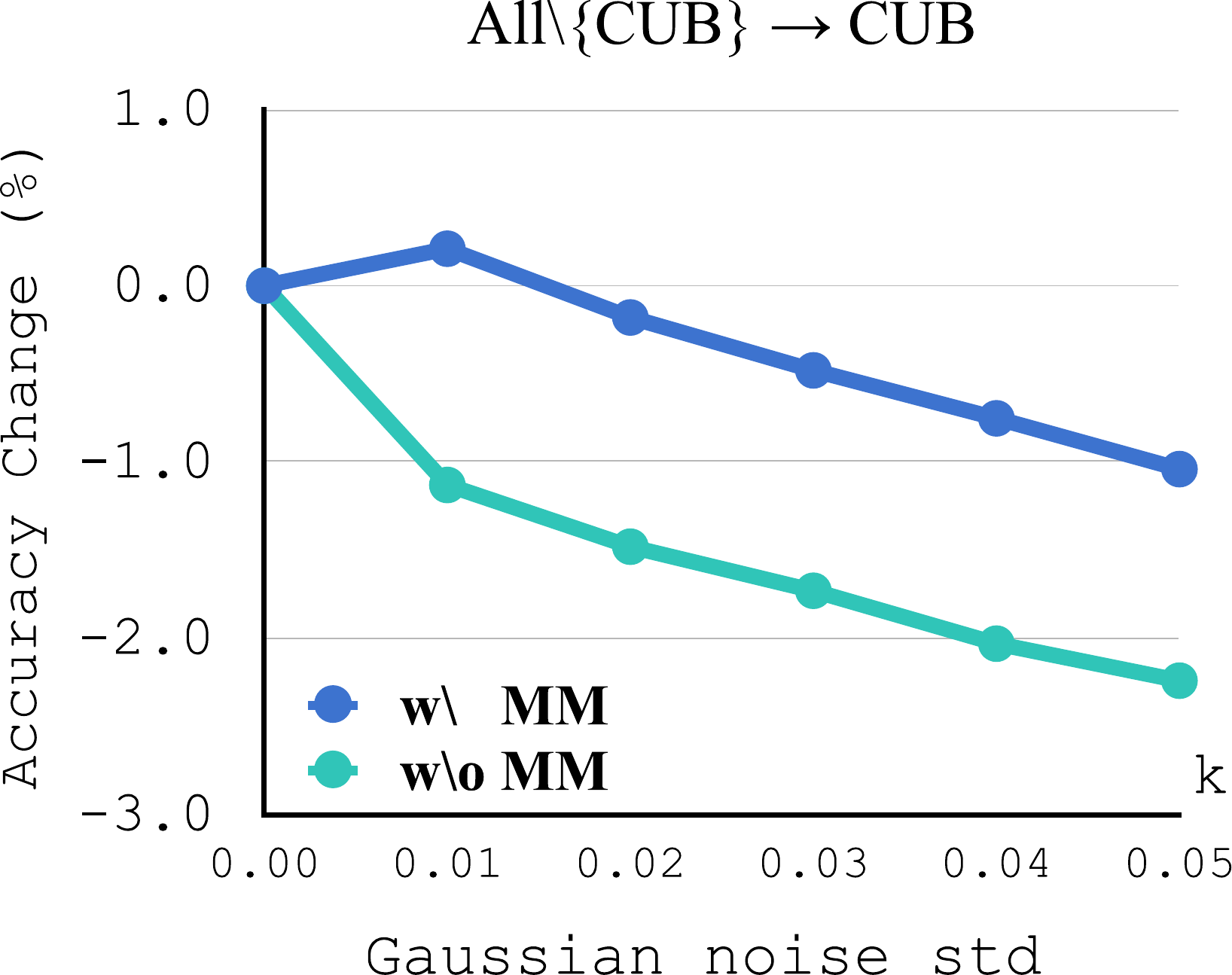}
    \includegraphics[width=0.245\textwidth]{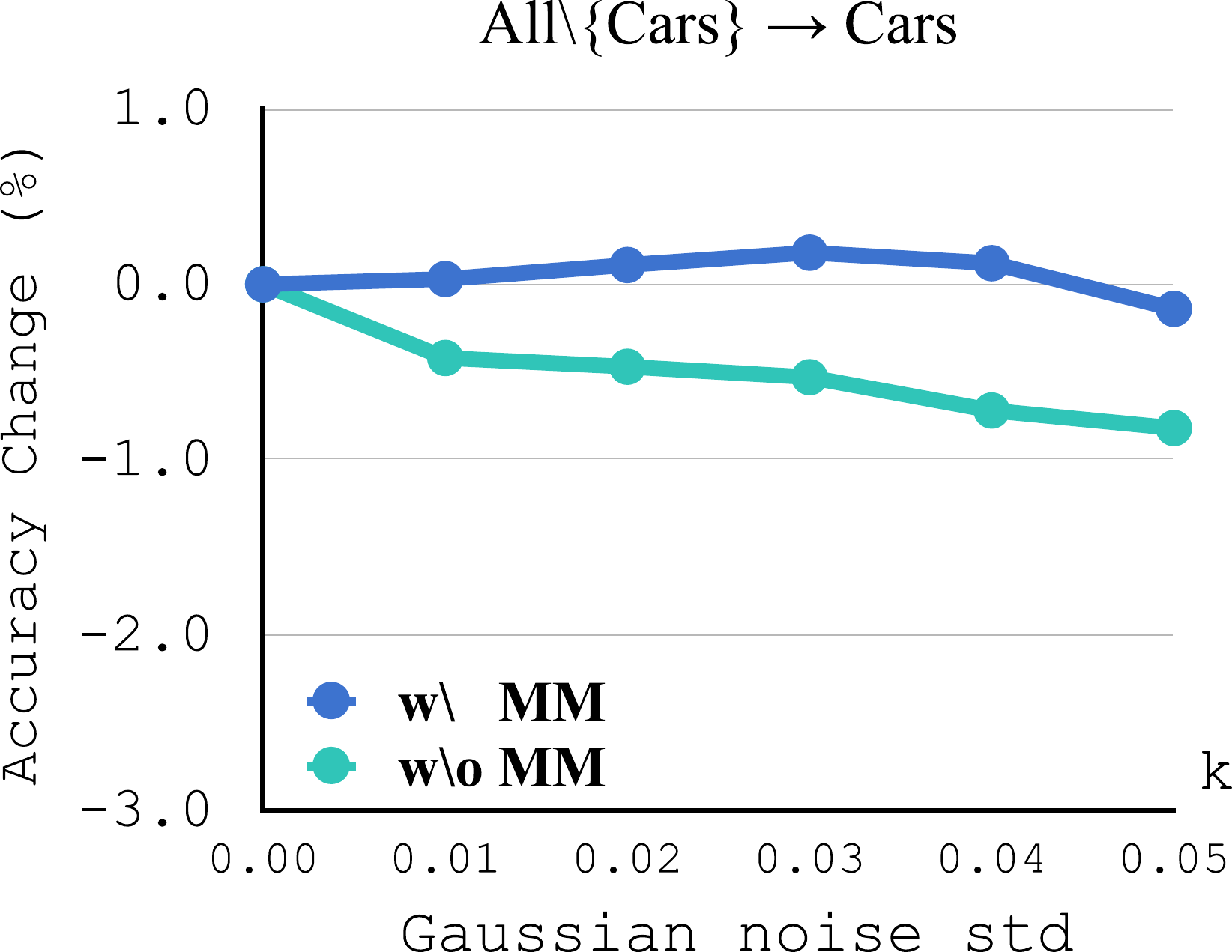}
    \includegraphics[width=0.245\textwidth]{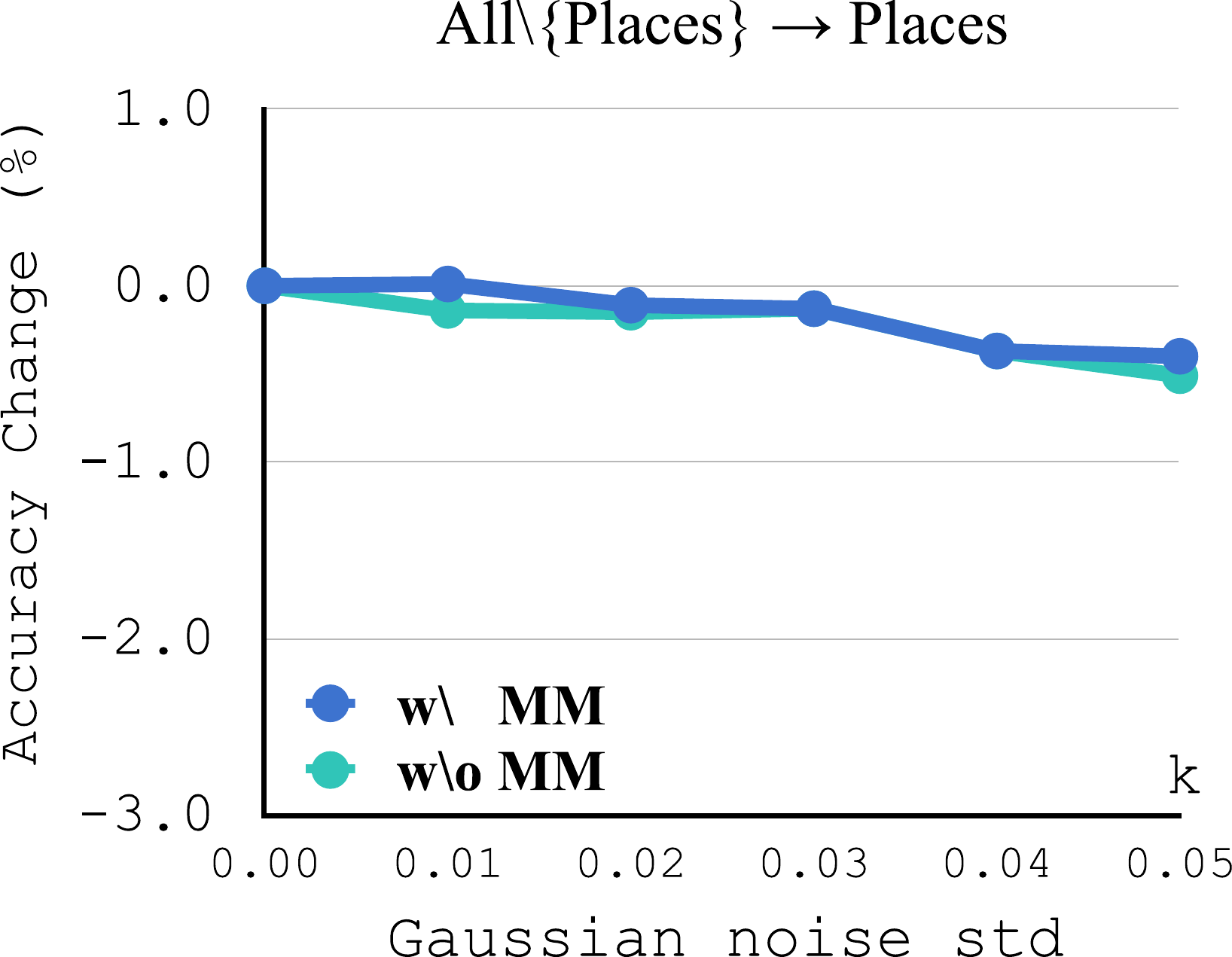}
    \includegraphics[width=0.245\textwidth]{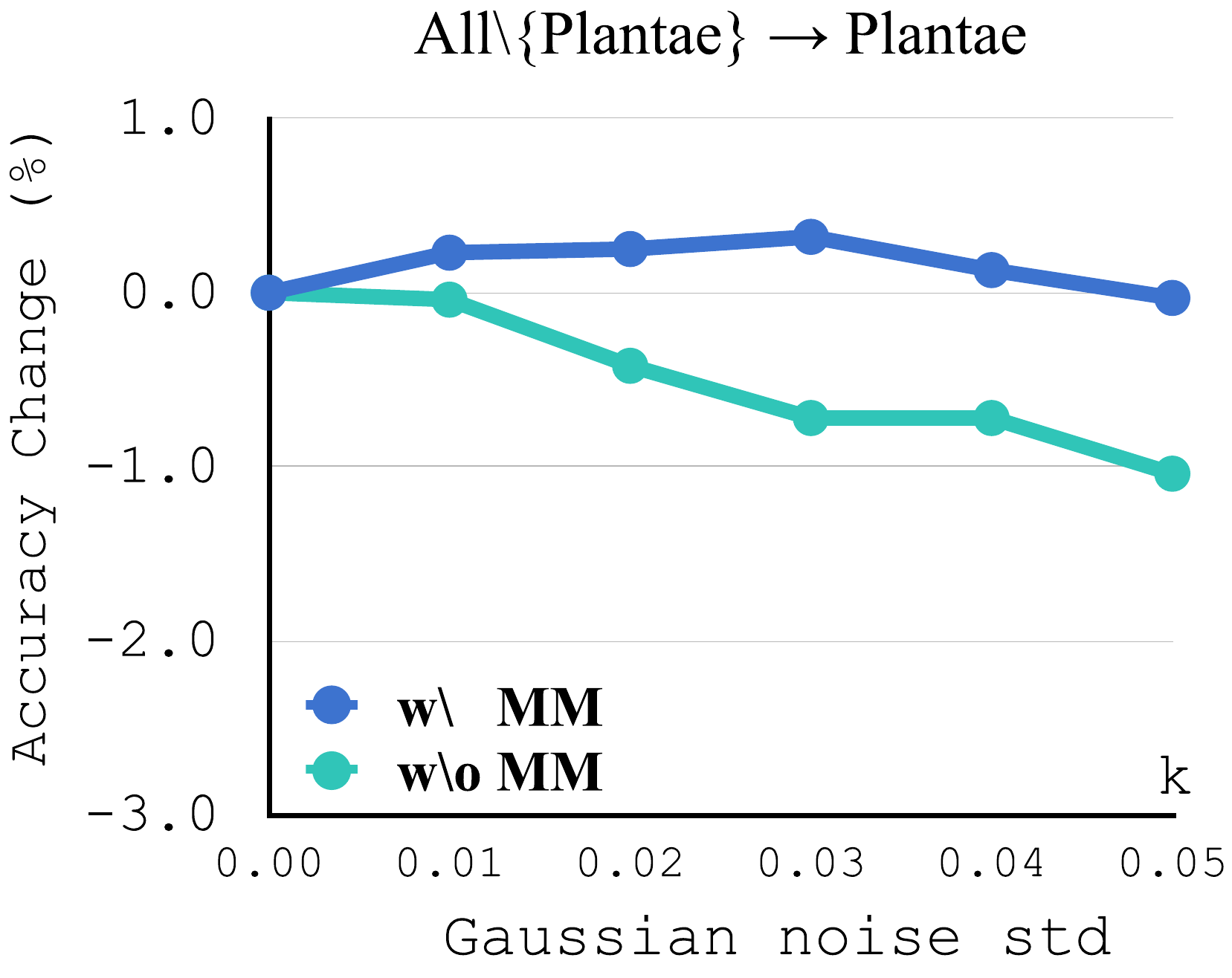}
    \caption{Minima quality analysis of $\mathcal{L}_{MM}$: Baseline FSC model vs. FS-BAN (only with $\mathcal{L}_{MM}$). We apply experiment setup i) with 5-Way 1-Shot tasks for training and testing, and observe the performance change by adding Gaussian noise with different std to all model parameters. We show that, compared to the baseline model, the domain robustness has been clearly improved.}
    \label{fig_robust}
 \end{figure*}
 \textbf{Mutual Regularization leads to better separation boundaries.} 
 In this analysis, we validate the effectiveness of $\mathcal{L}_{MR}$. For simplicity, models are trained on base classes of miniImageNet with 5-Way 1-Shot tasks.

 {In Figure \ref{mr_valid_acc}}, we visualize the performance of the teacher and the student in the meta-valid phase on tasks sampled from novel classes of miniImageNet. Compared to the baseline model and the original BAN (\ie, teacher without $\mathcal{L}_{MR}$), we observe that both the teacher and the student gain better generalization performance on novel classes. Meanwhile, the student can consistently outperform the improved teacher network, which suggests that $\mathcal{L}_{MR}$ maintains the advantage of the baseline BAN and brings non-degenerate solutions.
 
 Where does this improved performance come from?
 Qualitatively, in Figure \ref{lda_mr}, we follow \cite{gold-icml2020-unraveling} to sample tasks from novel classes (miniImageNet) and unseen domain (Places) and project the extracted features (via backbone network) of query samples onto the first two components of LDA \cite{mika1999fisher}, on directions that minimize the intra-class to inter-class variance ratio. In the plots, we observe that $\mathcal{L}_{MR}$ obtains better class separability, which leads to better generalization ability on novel classes of unseen domains. 

 Quantitatively, 
 we further follow \cite{gold-icml2020-unraveling}
 to analyze the quality of the learned features for few-shot tasks, via {feature clustering} (\(R_{FC}\)) and {hyperplane variation} (\(R_{HV}\)). 
 For measurement of \(R_{FC}\), we explicitly compute the intra-class to inter-class variance ratio.
 Denote the data in class 
 \(i\) and index \(j\) by \(\{x_{i, j}\}\), feature extractor by \(E\). \(\mu_{i}\) is the centroid feature of class \(i\) and \(\mu\) is the centroid feature of all classes, we have:
 \begin{equation}
    R_{FC}(E, \{x_{i, j}\}) =\frac{N_{w}}{N_{q}}\frac{\sum_{ij}\|E(x_{i,j})-\mu_{i}\|_2^2}{\sum_i \|\mu_{i}-\mu\|_2^2}, 
    \label{r_fc}
 \end{equation}
 where \(N_{w}\) and \(N_{q}\) are number of classes and query samples per class. 
 When \(R_{FC}\)=0, samples of the same category are mapped to a single point, and there is no uncertainty of hyperplane when separating arbitrary samples from two classes.  Similarly, Hyperplane Variation (\(R_{HV}\)) measures the sensitivity of separating hyperplanes to data sampling. For both \(R_{FC}\) and \(R_{HV}\), the lower value corresponds to better class separation.
 We compute \(R_{FC}\) and \(R_{HV}\) by sampling 200 query images per category, averaging over 1000 novel 5-Way 1-Shot tasks of the unseen domain. 
 These numerical results are shown in Table \ref{table5}. Furthermore, as TSD analysis in Sec. VI.G, it is clear that the improvement comes from the awareness of cross-category information of the teacher network, thus a simple $\mathcal{L}_{MR}$ brings better class separation and feature clustering performance on unseen domains.
 
 \textbf{Temperature convergence and analysis with $\mathcal{L}_{MCT}$.} We visualize the meta-learned temperature (initialized by $\tau$=4) trained with experiment setup i), as {Figure \ref{fig_temperature}}.
 In both 1-Shot and 5-Shot training processes, the meta-controlled temperature gradually converges, finding its own equilibrium. Therefore, the adaptively tuned hyperparameter on diverse domains is the reason that we obtain the improvements, as numerical results in Table \ref{table_ablation_study}. Meanwhile, in Table \ref{mct_direct}, we further note that directly updating the temperature as a learnable parameter does not bring improvement and introduces overfitting. 
 
 \textbf{Mismatched Teachers improve solution robustness.} 
 How to understand that a randomly selected and mismatched teacher of FS-BAN improves robustness to DG-FSC (see Table \ref{table2})? One ideal case is that converging to a `wide' minima leads to a more robust solution of the model. Recently, some DG literature on conventional supervised learning \cite{chaudhari2019wide-valley, li2019episodic_random, keskar2016sharp} analyze the model robustness in terms of evaluating the solution minima quality. 
 
 Following \cite{li2019episodic_random, keskar2016sharp}, we compare the model robustness, by adding Gaussian noise to model parameters and observe the accuracy change in the testing phase, as {Figure \ref{fig_robust}}. In most cases, we can observe that FS-BAN (with only $\mathcal{L}_{MM}$) brings higher robustness facing perturbation, which suggests better minima quality and generalization on held-out unseen domains. Another interesting observation is that in some cases we obtain an incremental improvement by introducing noise to model weights, which is a by-product that is related to a recent work \cite{tseng2020cross}.

 %%%%%%
 %%%%%% TSD
 %%%%%%
 \subsection{Top-score Difference Analysis}
 \label{sec_tsd}
 % Top-score difference
 In their previous work \cite{yang2019aaai-second}, they show that a better student model can be obtained with a more tolerant teacher, which is less focused on the primary class when making predictions. That is, the teacher passes the reasonable inter-class knowledge to the student (\ie, probability prediction to all categories). Following their findings, in episodic training, we measure the Top-score difference (TSD) of the probability predictions produced by the teacher network:
 \begin{equation}
     \textnormal{TSD}=f_{\theta_0, a1}(\cdot) - \frac{1}{{M}-1}\sum_{m=2}^{M}f_{\theta_0, am}(\cdot),
 \end{equation}
 where \(f_{\theta_0, am}(\cdot)\) is short for \(m\)-th largest value in the probability distribution \(f_{\theta_0}(\cdot)\). We set a fixed \(M=3\) which represents the number of potential semantically similar classes for each image in the episode, including the primary class (the class assigned the highest probability). Then, we calculate the gap between the prediction probabilities of the primary class and the average of other \(M-1\) classes with the highest scores.
 %%%%%%%%%% %%%%%%%%%%%%%%%%%
 % ablation study of FS-BAN
\begin{table*}[t]
    \centering
    \caption{
    {
    % \color{blue}
    We compare the meta-test accuracy (\%) on unseen domains to the method proposed by Tian \etal \cite{tian2020rethinkingsd-fsc}. Model is trained on several seen source domains and evaluated on the leave-one-out selected unseen domain with 5-Way 5-Shot tasks.}}
    \label{table_ablation_tian_baseline}
    % \begin{adjustbox}{width=0.99\textwidth}
    {
    % \color{blue}
    \begin{tabular}{l cccccccc}
        \toprule
             \textbf{Method} & {{All$\backslash$ \{CUB\}$\mapsto$CUB}}  & {{All$\backslash$ \{Cars\}$\mapsto$Cars}}  & {{All$\backslash$ \{Places\}$\mapsto$Places}}  & {{All$\backslash$ \{Plantae\}$\mapsto$Plantae}}  \\
            \hline
            
            [43] (tian \etal \cite{tian2020rethinkingsd-fsc}) & $56.07 \pm 0.77$ & $41.22 \pm 0.59$ & $67.73 \pm 0.61$ & $52.97 \pm 0.50$\\\hline
            
            MatchingNet \cite{vinyals2016matching} & \(51.92 \pm 0.80\) & \(39.87 \pm 0.51\) & \(61.82 \pm 0.57\) & \(47.29 \pm 0.51\) \\
            
            + FS-BAN (Ours) & \bm{$61.34 \pm 0.52$} & \bm{$45.01 \pm 0.57$} & \bm{$70.09 \pm 0.60$} & \bm{$53.89 \pm 0.64$} \\\hline

            RelationNet \cite{sung2018relationnet} & \(62.13 \pm 0.74\) & \(40.64 \pm 0.54\) & \(64.34 \pm 0.57\) & \(46.29 \pm 0.56\)\\
            
            + FS-BAN (Ours) & \bm{$65.55 \pm 0.56$} & \bm{$45.78 \pm 0.57$} & \bm{$69.72 \pm 0.58$} & \bm{$53.55 \pm 0.57$} \\ \hline

            GNN \cite{garcia2017gnn} &
            \(69.26 \pm 0.68\) & \(48.91 \pm 0.67\) & \(72.59 \pm 0.67\) & \(58.36 \pm 0.68\)\\
            
            + FS-BAN (Ours) & \bm{$73.70 \pm 0.66$} & \bm{$50.66 \pm 0.65$} & \bm{$78.57 \pm 0.67$} & \bm{$61.85 \pm 0.66$} \\
        \bottomrule
    \end{tabular}}
    \vspace{-4 mm}
    % \end{adjustbox}
    \end{table*}

 \textbf{TSD for $\mathcal{L}_{MR}$}.  In FS-BAN, $\mathcal{L}_{MR}$ requires the teacher network, \ie \(f_{\theta_0}(\cdot)\), to match the soft distribution produced from the student, \ie \(f_{\theta_1}(\cdot)\). Therefore, the teacher can learn the cross-category similarity information from the student. Here, we quantify these benefits via statistical measurements during training. 
 As shown in Table \ref{table-a3}, in the training process, $\mathcal{L}_{MR}$ indirectly reduces TSD of the teacher network, which suggests that the produced soft predictions are less picked and the similarity knowledge is well preserved. In the testing phase, $\mathcal{L}_{MR}$ for FS-BAN has higher accuracy. Therefore, the teacher network is less overfitting and preserves the meaningful soft knowledge transferred from the student.

 \textbf{TSD for $\mathcal{L}_{MM}$}. 
 What does the student learn from the mismatched teacher? To understand the working mechanism of FS-BAN mismatched teachers, a potentially ideal way is to observe the behavior of the mismatched teacher. We select the different source domains, then we observe the performance of the teacher training on the miniImageNet (hence, when miniImageNet is not the source domain, the teacher becomes a mismatched teacher). We sample 5-Way 5-Shot tasks from the novel classes of each domain, and we measure the TSD and the accuracy of the teacher.  As Table \ref{Table-MM}, when we evaluate the teacher network on a mismatched source domain, we find that the accuracy is far beyond the random prediction. Therefore, the mismatched teacher is at least meaningful since it is better than randomly guessing. On the other hand, it has apparently lower TSD compared to that of miniImageNet in meta-testing. In this DG-FSC scenario, the attention of the mismatched teacher has transited to predicting the inter-class similarity, and the student is trained to adapt unseen domain by adapting to the ``unseen" (mismatched) teacher. At the same time, the student model can be optimized by cross-entropy loss to the ground truth, which guarantees its correct updating directions.

 {
 % \color{blue}
 In literature, to improve the model generalizability of unseen samples for classification tasks, several regulators such as Label Smoothing \cite{szegedy2016labelsmoothing} or Confidence Penalty \cite{pereyra2017confidencepenalty} have been proposed to penalize the overconfidence prediction of the classifier, such that the overfitting for the training data is mitigated.
 However, we note that these regulators have a common drawback: they encourage the probabilities to be uniformly distributed over all training classes, regardless if these classes are really similar to each other.  
 % For example, in Table \ref{table_r1_lmr_lmm_ablation}, we observe that the proposed $\mathcal{L}_{MM}$ performs better than a noisy teacher model, which is randomly initialized, similar to the label smoothing technique \cite{szegedy2016labelsmoothing}.
 In contrast to this, in our proposed method, the student in $\mathcal{L}_{MR}$ and $\mathcal{L}_{MM}$ is regularized to match a soft and better confidence prediction, which is designed specifically for overfitting and domain-shift for DG-FSC, and they achieve considerable improvement.
 }

 {
 % \color{blue}
 \subsection{Training Student with a Stronger Teacher}
 % In this work, our proposed method is inspired by born-again networks \cite{furlanello2018bornagain} in conventional supervised learning, where the student has the same network architecture as the teacher network. 
 In Sec. \ref{sec4}, to find a good balance between the performance and the training cost, the proposed FS-BAN does not involve sequential training in generations, and we only train one generation of the student. 
 Therefore, the architecture and size of the student are not limited to being the same as the teacher.
 
 Ideally, one possible way to further improve the performance of the student network is to introduce a stronger teacher network, \ie, more parameters with higher capacity. 
 In Table \ref{table_larger_backbone}, we conduct a study to empirically validate this assumption: 
 We set the scale of the teacher backbone equal (\textit{born-again networks setup}) or larger (\textit{common knowledge distillation setup}) than that of the student.
 We consider different types of backbone networks that are popular in FSC \cite{tian2020rethinkingsd-fsc, tseng2020cross, finn2017maml, snell2017prototypical, sung2018relationnet} as the feature encoder: Conv-4/6 (4/6-layer convolutional networks), and ResNet-10/18 \cite{he2016resnet}.
 We use the same setting as experiment setup 1): the student is trained on multiple seen source domains and tested on the leave-one-out selected target domain with 5-Way 5-Shot tasks. We use MatchingNet \cite{vinyals2016matching} as the baseline model.
 
 As can be observed in Table \ref{table_larger_backbone}, when the backbone networks of the teacher and the student are the same, our proposed FS-BAN improves the performance of the student by a considerable margin. 
 On the other hand, if we choose a teacher network with a larger backbone, the  performance of the student network can be further improved.
 
 \subsection{Comparison to BAN with Transfer Learning}
 In this section, we compare our proposed method with the simple baseline \cite{tian2020rethinkingsd-fsc} that leverages BAN with transfer learning approach for FSC, using the experiment setup 1), where we have multiple source domains: for each seen source domain, we follow \cite{tian2020rethinkingsd-fsc} to initialize a linear layer as the classifier head and they share the feature encoder (ResNet10 \cite{he2016resnet}). 
 In each training epoch, we randomly select the source domain and the corresponding classifier head, and the model is optimized by minimizing a standard cross-entropy loss as Eqn. \ref{eq1}. 
 In DG-FSC evaluation, we follow \cite{tian2020rethinkingsd-fsc} to transfer the obtained feature encoder on novel tasks and fit a new linear classifier for the prediction of query samples. 
 For a fair comparison, we apply the same backbone and data augmentation skills as our method. 
 The results are in Table \ref{table_ablation_tian_baseline}. 
 We show that our proposed method can achieve competitive performance on different DG-FSC setups. Moreover, we note that we follow \cite{tian2020rethinkingsd-fsc} to conduct BAN training for two generations, therefore their training cost is higher than that of our method.
 }

% section 7 conclusion

 \section{Discussion}
 \label{sec6}
\textbf{Conclusion.}
 In this work, we first propose Born-Again Network (BAN) episodic training for domain generalization few-shot classification (DG-FSC) and reveal that BAN leads to more discriminative features and generates better decision boundaries on novel tasks from unseen domains. 
 This suggests that similar to the observation in conventional supervised learning, BAN is also promising for DG-FSC tasks.
 To the best of our knowledge, this is the first study of BAN for episodic training.
 Motivated by this, we propose Few-Shot BAN (FS-BAN) as our main contribution.
 FS-BAN consists of multi-task learning objectives: Mutual Regularization, Mismatched Teacher, and Meta-Control of the Temperature. They aim to address the unique challenges posted specifically in DG-FSC: overfitting and domain shift. The effectiveness of FS-BAN is demonstrated by competitive accuracy on six benchmark datasets, three baseline FSC models, and qualitative and quantitative ablation studies. 

\textbf{Limitation.}
 We follow exactly previous work (\eg, \cite{tseng2020cross}) in the choice of domains and datasets for a fair comparison. 
 However, given the extremely wide range of domains to which DG-FSC can be applied, it is not feasible for us to validate our findings for all possible domains.
 On the other hand, our comprehensive qualitative and quantitative experiment results supported by our analysis provide supportive evidence that our method could be generalized to other domains.
 Meanwhile, FS-BAN does not impact the inference stage since we do not modify the model structure. Therefore, the effectiveness of FS-BAN on other domains in the open world can be easily validated with existing FSC models.
 
\textbf{Future Work.}
 While the performance of the state-of-the-art FSC algorithms has been largely improved within the single domain and unseen domains that include diverse classes, the accuracy on fine-grained domains remains poor, and an example can be observed in the results of the Cars domain in Table \ref{table2}. Future work will consider the different types of unseen domains, including this fine-grained setup that is challenging for all current FSC models.

% \appendices

% or



% \section{}

% \section*{Acknowledgment}


\ifCLASSOPTIONcaptionsoff
  \newpage
\fi


\bibliography{IEEE.bib}{}
\bibliographystyle{plain}

{
% \small
\section*{Acknowledgment}
{
% AISG
This work was supported in part by the National Research Foundation, Singapore, under its AI Singapore Programmes (AISG) under Award AISG2-RP-2021-021 and Award AISG2-TC-2022-007; 
% SGP
and in part by the Singapore University of Technology and Design under Project PIE-SGP-AI-2018-01.
% HTIF
This project was also based on the research/work support in part by the Changi General Hospital
and Singapore University of Technology and Design, under the HealthTech Innovation Fund (HTIF Award No. CGH-SUTD-2021-004).
% others
The authors would like to thank anonymous reviewers for their helpful comments to improve the paper and also would like to thank Yiluan Guo, Jiamei Sun and Milad Abdollahzadeh for their valuable comments, feedback and discussion.
}
}

\end{document}